%% file: main.tex
\documentclass[pmlr,twocolumn,10pt]{jmlr} %

\usepackage{booktabs}
\usepackage{siunitx}

\usepackage[switch]{lineno}

\usepackage{longtable}
\usepackage{makecell}
\usepackage{algorithm}

\usepackage{wrapfig}
\usepackage{adjustbox}
\usepackage[most]{tcolorbox}
\usepackage{xcolor}
\usepackage{booktabs}
\usepackage{ragged2e}
\usepackage{array}
\newcolumntype{P}[1]{>{\RaggedRight\arraybackslash}p{#1}}

\tcbset{
  annopanel/.style={
    enhanced,
    colback=gray!4,
    colframe=gray!35,
    boxrule=0.4pt,
    arc=2mm,
    left=3mm,right=3mm,top=2mm,bottom=2mm,
  },
  annobox/.style={
    enhanced,
    colback=white,
    colframe=gray!120,
    boxrule=0.4pt,
    arc=1.5mm,
    left=2mm,right=2mm,top=1.5mm,bottom=1.5mm,
    fonttitle=\footnotesize\bfseries,
  }
}

\newcommand{\bftab}{\fontseries{b}\selectfont}

\newcommand{\annottablestyle}{%
  \scriptsize
  \setlength{\tabcolsep}{5pt}%
  \renewcommand{\arraystretch}{1.12}%
}

\definecolor{darkpurple}{RGB}{64, 0, 128} %
\definecolor{darkred}{RGB}{139, 0, 0}     %
\definecolor{darkteal}{RGB}{0, 102, 102}  %

\theorembodyfont{\upshape}
\theoremheaderfont{\scshape}
\theorempostheader{:}
\theoremsep{\newline}

\jmlrvolume{333}
\jmlryear{2026}
\jmlrsubmitted{LEAVE UNSET}
\jmlrpublished{LEAVE UNSET}
\jmlrworkshop{Conference on Health, Inference, and Learning (CHIL) 2026} %

 \title[Reconstructing Sepsis Trajectories from Clinical Case Reports]{Reconstructing Sepsis Trajectories from Clinical Case Reports using LLMs: the Textual Time Series Corpus for Sepsis}

\author{%
\Name{Shahriar Noroozizadeh} \Email{snoroozi@cmu.edu}\\
\addr Machine Learning Department and Heinz College\\
Carnegie Mellon University, USA 
\AND
\Name{Jeremy C. Weiss} \Email{jeremy.weiss@nih.gov}\\
\addr National Library of Medicine\\ 
National Institutes of Health, USA 
\vspace{-2.5em}
}

\begin{document}

\maketitle

\begin{abstract}
Clinical case reports and discharge summaries may be the most complete and accurate summarization of patient encounters, yet they are finalized, \emph{i.e.}, timestamped after the encounter.
Complementary structured data streams become available sooner but suffer from incompleteness.
To train models and algorithms on more complete and temporally fine-grained data, we construct a pipeline to phenotype, extract, and annotate time-localized findings within case reports using large language models.
We apply our pipeline to generate an open-access textual time series corpus for Sepsis-3 comprising 2,139 case reports from the PubMed-Open Access (PMOA) Subset.
To validate our system, we apply it to PMOA and timeline annotations from i2b2/MIMIC-IV and compare the results to physician-expert annotations.
We show high recovery rates of clinical findings (event match rates: GPT-5--0.93, Llama 3.3 70B Instruct--0.76) and strong temporal ordering (concordance: GPT-5--0.965, Llama 3.3 70B Instruct--0.908). 
Our work characterizes the ability of LLMs to time-localize clinical findings in text, illustrating the limitations of LLM use for temporal reconstruction and providing several potential avenues of improvement via multimodal integration.
\end{abstract}

\vspace{-0.2em}
\paragraph*{Data and Code Availability}
This paper uses the i2b2 \citep{sun2013evaluating}, MIMIC-IV \citep{johnson2023mimic}, and PubMed Open Access (PMOA) datasets \citep{pmc2024open}, as well as the timeline annotations from \cite{frattallone2024using}.
Code and annotations are available at: \url{https://github.com/Shahriarnz14/T2S2}.

\paragraph*{Institutional Review Board (IRB)}
This research has been designated by our IRB as Not Human Subjects Research.

\section{Introduction}
\label{sec:intro}

The Third International Consensus Definitions for Sepsis (Sepsis-3) are the benchmark definitions for sepsis, the dysfunctional immune response to an infection, which are used in trial eligibility criteria \citep{kyriazopoulou2021procalcitonin} and phenotyping studies \citep{seymour2019derivation}.
In the critical care literature, sepsis is seen as a heterogeneous disease, a pathway arrived upon from a multitude of infectious origins, anatomical sites, and comorbid profiles.  
There is increasing recognition of the importance of time in understanding sepsis and sepsis progression, as seen in early-warning systems \citep{henry2022factors}, variation in prediction utility \citep{kamran2024evaluation}, and endotype characterization \citep{noroozizadeh2023temporal}.

Meanwhile, structured data streams, which are the data resources these tools use, are often incomplete records with insufficient information to render or confirm a diagnosis of sepsis \citep{moldwin2021empirical}. %
A more complete alternative is the discharge summary, or more generally, the case report, in which a clinician documents the clinical findings pertinent to the case and care management.
The impracticality of using the case report is that it is only written in full after the encounter is over, \emph{i.e.}, with a less granular timestamp than the events contained within. 
In principle, having a more complete, fine-grained temporal record could enable better temporal analyses of patients with sepsis, leading to better understanding and treatment improvements.

At present, the (semi-)public critical care repositories at the research frontier contain high resolution structured signals and low resolution textual signals. 
While several concerted efforts have focused on extracting temporal information, they have largely focused on temporal relations between clinical concepts \citep{sun2013evaluating} rather than on event timing.  
Those that have focused on timing \citep{leeuwenberg2020towards,frattallone2024using} used excerpted summaries and relatively small sample sizes.

In order to train sepsis models that can incorporate information referenced in free text with finer temporal granularity, we seek to generate larger samples of textual time series.
To do so, we introduce a novel Sepsis-3 textual time series corpus from the PubMed Open Access (PMOA) Subset.
We adopt an LLM-as-annotator approach, evaluating both the ability of models to extract clinical findings and their ability to associate those findings with timestamps.
We augment existing clinical annotations with our own PMOA annotations and contrast LLM annotations with both our clinician annotations and the annotations of \cite{frattallone2024using}.
Because the corpus is automatically generated beyond the manually annotated subset, we validate it through a multi-layer framework: clinician-annotated gold-standard evaluation on \texttt{sepsis-40}, clinician--clinician agreement, bootstrap uncertainty quantification, bronze-standard proxy-reference evaluation on \texttt{sepsis-100}, direct hallucination auditing, multi-pass self-consistency, and full-corpus cross-model agreement.
These assessments provide evidence for the quality and limitations of the LLM annotations and support the use of T2S2 as a retrospective research resource for temporal modeling of sepsis trajectories.

\begin{figure*}[th]
\begin{minipage}[t]{\dimexpr \textwidth\relax}
\hspace{2em}
  \begin{minipage}[t]{0.65\textwidth}
    \vspace{8pt}  %
    \includegraphics[width=\linewidth]{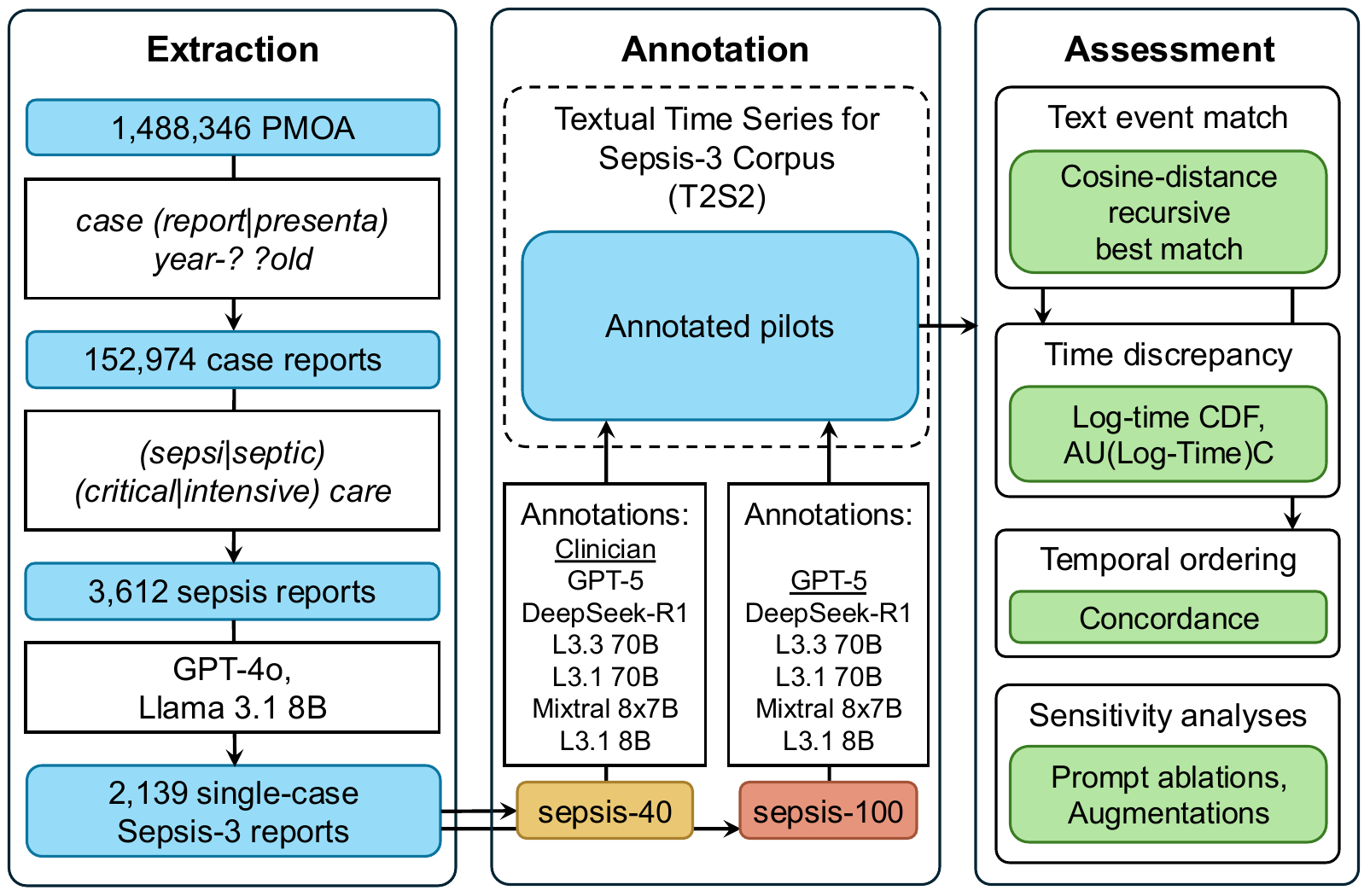}
  \end{minipage}%
  \hspace{.5em}
  \begin{minipage}[t]{\textwidth}
    \vspace{1em}  %
    \begin{tabular}{r|cc}
    \multicolumn{3}{c}{\textbf{i2m4}} \\
    \midrule
    Qwen\textbackslash Human & Yes & No \\
    Yes & 2 & 0 \\
    No  & 0 & 18 \\
    \toprule
    L3.1 8B & Yes & No \\
    Yes & 2 & 7 \\
    No  & 0 & 11 \\
    \midrule
    \multicolumn{3}{c}{\textbf{\texttt{sepsis-100}}} \\
    \toprule
    GPT-4o\textbackslash Qwen & Yes & No \\
    Yes & 78 & 7 \\
    No  & 8 & 7 \\
    \toprule
    L3.1 8B & Yes & No \\
    Yes & 81 & 12 \\
    No  & 5 & 2 \\
    \bottomrule
    \end{tabular}
  \end{minipage}

  \vspace{0.3em}
  \caption{Flowchart of the PMOA T2S2 corpus construction pipeline (left).
    Confusion matrices (right) for Sepsis-3 phenotype in i2m4 (top) and \texttt{sepsis-100} (bottom) datasets, with preferred annotator in column position.}
  \label{fig:flow}
  \vspace{-2em}
\end{minipage}
\end{figure*}

\begin{sloppypar}
\subsection{Related work}
Numerous critical care and sepsis datasets are available \mbox{(semi-)publicly}, \emph{e.g.}, \cite{thoral2021sharing,pollard2018eicu}, including several with de-identified clinical discharge summaries: MIMIC-III \citep{johnson2016mimic}
and MIMIC-IV \citep{johnson2023mimic}. 
Sepsis phenotyping (including Sepsis-3) has been conducted using a variety of methods on structured data \citep{johnson2018comparative}, although structured data is often highly missing, leading to suboptimal phenotyping \citep{seinen2025using}.
Additionally, the timing of sepsis phenotyping is crucial, in that delays in sepsis prediction can nullify predictiveness and lead to burdensome and unhelpful analytic models \citep{kamran2024evaluation}.  
\end{sloppypar}

Several works have used the i2b2 (a competition subset of MIMIC-II/III) to construct timelines of clinical concepts \citep{leeuwenberg2020towards,frattallone2024using}.  
In the case of \cite{leeuwenberg2020towards}, the event spans are already given, and focus on an excerpt of the full discharge summary, so we focus our direct comparison on that of \cite{frattallone2024using}.
In contrast to these works, we adopt an annotation process focused on temporal assignment to clinical findings rather than i2b2 clinical concepts \citep{uzuner20112010}, which allows for greater specificity of the event in the absence of additional context.  
Additionally, as compared to \cite{frattallone2024using},
we focus on using text alone for clinical time series construction, which is necessary for the PubMed Open Access Subset since no complementary data source is available.

Broadly in the medical domain, LLMs are showing promise in medical tasks such as summarization \citep{van2024adapted}.
In fact, some findings suggest that 
medical fine-tuning rarely leads to improved performance over the corresponding foundation models \citep{jeong-etal-2024-medical}. 
In our analysis, we primarily use foundation LLMs and show that their scale and instruction-tuning are central to the performance of the model.  

\subsection{Contributions}
Our work highlights the following contributions:
(1) we present the first textual time series corpus for sepsis, comprising 2,139 open-access case reports identified as having Sepsis-3;
(2) we develop an analytic pipeline to identify Sepsis-3 from clinical text and to extract time-associated clinical findings;
(3) we contribute clinician annotations for 40 PMOA case reports, yielding $3,229$ manually annotated clinical reference events;
(4) we develop and adapt evaluation criteria for textual time series, where matched clinical findings are not expected to be identical text strings;
(5) we evaluate extraction performance across multiple datasets, model families, prompt variants, and time scales; and
(6) we provide a multi-layer validation framework for the automatically generated corpus, including bootstrap uncertainty estimates, clinician--clinician agreement, hallucination auditing, self-consistency analysis, and full-corpus cross-model agreement.

This work goes beyond simple event extraction by establishing textual time series as a distinct representation for clinical narratives.
We explicitly address the misalignment between narrative structure and chronological reality---where retrospective summaries often reveal outcomes before precursors, inducing causal leakage during predictive modeling.
By validating this representation through a framework that measures semantic content recovery, temporal ordering, time-scale-specific timestamp accuracy, factual grounding, and annotation stability, we provide a reusable retrospective resource for downstream temporal modeling tasks such as forecasting, causal reasoning, and multimodal alignment.

\section{Methods}
In this section, we first formalize the problem of extracting textual time series from clinical narratives and detail the multi-stage pipeline used to construct the T2S2 corpus. We then outline our evaluation methodology, which combines embedding-based event matching, temporal alignment metrics, and expert physician review.

\vspace{-1em}
\subsection{Task Formalization}
Given a clinical text document $T$, we extract a textual time series $S = \{(e_1,t_1), (e_2,t_2),...,(e_n,t_n)\}$ where $e_i$ represents a clinical finding (contiguous text span describing a temporally localizable patient event) and $t_i \in \mathbb{R}$ represents hours relative to admission ($t=0$).

We define clinical findings with two key modifications from i2b2 clinical concepts \citep{uzuner20112010}: (1) findings extend beyond single prepositional phrases for specificity (e.g., ``pain in chest that radiates substernally'' remains intact), and (2) conjunctions split into components (e.g., ``metastases in liver and pancreas'' $\rightarrow$ \{``metastasis in liver'', ``metastasis in pancreas''\}). Temporal assignment captures event start times (e.g., ``three-day history of fever'' $\rightarrow$ $t=-72$).

For evaluation, given predicted $S_{pred}$ and reference $S_{ref}$, we compute: event matching via cosine distance of embeddings (threshold $\theta=0.1$; sensitivity analysis in Appendix~\ref{apd:threshold_sweep}), temporal ordering via concordance, and time accuracy via area under log-time curve (AULTC; see Appendix \ref{apd:ltcdf}).

\subsection{Data and Extraction Pipeline}
We developed a four-stage pipeline (Figure \ref{fig:flow}) for case report detection, Sepsis-3 phenotyping, clinical finding extraction, and timestamp determination, applied to two data sources: (1) the absolute timeline annotations from \cite{frattallone2024using}, and (2) the PubMed Open Access subset \citep{pmc2024open} (see Table \ref{tab:dataset-summary}).

\input{tables/datasets.tex}

\subsubsection{Datasets}
\label{section:datasets}

\paragraph{i2m4.} This dataset contains 20 discharge summaries (15 i2b2, 5 MIMIC-IV) with physician-annotated temporal concepts from \cite{frattallone2024using}. We used interval lower bounds as event times and only probabilistic annotations. For Sepsis-3 phenotyping, we used a prompt describing qSOFA criteria: suspected/confirmed infection with blood pressure/respiratory rate/mental status abnormalities, requesting binary output (1 for sepsis, 0 otherwise).
To develop a computational approach, we used a prompt describing qSOFA criteria (Appendix \ref{apd:phenotype-prompt}) with QwQ-32B-Preview achieving 100\% accuracy and Llama-3.1-8B-Instruct achieving 65\% accuracy.

\paragraph{PMOA (\texttt{sepsis-40}, \texttt{sepsis-10}, \texttt{sepsis-100}).}
From 1.5M publications (version April 25, 2024), we identified case reports via regex (\texttt{case (report|presenta)} and \texttt{year-?old}), then filtered for sepsis using \texttt{(sepsi|septic)} and \texttt{(critical|intensive) care}.
GPT-4o and Llama-3.1-8B screened for Sepsis-3, yielding 2,139 cases.
We manually annotated 40 PMOA case reports for the gold-standard \texttt{sepsis-40} evaluation set, yielding $3,229$ clinician-annotated reference events.
The \texttt{sepsis-10} subset is a 10-report subset of \texttt{sepsis-40} used for clinician annotation quality review, hallucination audit, multi-pass self-consistency, and sepsis onset analysis.
We also sampled 100 non-overlapping PMOA case reports for \texttt{sepsis-100}, which is used as a bronze-standard proxy-reference set annotated by GPT-5 to evaluate extraction behavior at larger scale rather than as clinical ground truth.
The full case identifiers for \texttt{sepsis-40} and \texttt{sepsis-10} are provided in Appendix~\ref{apd:subset_ids}.
Figure~\ref{fig:flow} (right) shows phenotyping confusion matrices.

\textbf{Phenotyping output parsing and clinician audit.}
In practice, some LLMs produce verbose rationales rather than a minimal binary output. We therefore parsed phenotyping decisions using a rule-based extraction step that searches model responses for explicit affirmative/negative indicators of meeting sepsis criteria and maps them to a binary label. If primary extraction did not yield a valid label, a fallback strategy examined the final line of the response, and discordant parses were flagged for manual inspection. We additionally performed clinician audit on sampled and borderline cases to verify alignment with Sepsis-3 intent under information available in narratives.

\subsubsection{Clinical Finding Extraction and Temporal Assignment}
We queried six LLMs (GPT-5, DeepSeek-R1-UD-IQ1\_S, Llama-3.3-70B-Instruct, Llama-3.1-70B-Instruct, Mixtral-8x7B-Instruct v0.1, Llama-3.1-8B-Instruct) using a structured prompt with few-shot examples to extract event-time pairs. The prompt instructs models to: assign admission time as $t=0$, use negative times for pre-admission events, split conjunctions, and capture event start times for durations (full prompt in Appendix \ref{apd:prompt}). For sensitivity analysis, we tested ablations (no-role, zero-shot, no-expansion) and augmentations (interval bounds, i2b2 typing).

For i2m4, we compared against a BERT baseline adapted from \cite{frattallone2024using}, using a 3-layer MLP regressor with L1 loss for time prediction. We filtered events to $\pm$2 weeks of admission for training stability (3,472$\rightarrow$3,350 findings) and used 5-fold cross-validation.

\subsection{Evaluation Methodology}
\subsubsection{Quantitative evaluation of extracted timelines}
\label{sec:evaluation_methodology_metrics}
We evaluate extracted textual time series using four metrics:\\
\textbf{Event match rate}: Proportion of reference events matched via recursive best matching with cosine distance threshold 0.1 using PubMedBert embeddings (Appendix~\ref{apd:pseudo}).\\
\textbf{Concordance (c-index)}: Probability of correctly ordering random event pairs.\\
\textbf{Median absolute error (MAE)}: Median time discrepancy (in hours) among matched events.\\
\textbf{Area under log-time curve (AULTC)}: Integrated log-transformed discrepancies capturing accuracy across time scales (detailed derivation in Appendix \ref{apd:ltcdf}).

A clinician qualitative review of annotation quality on \texttt{sepsis-10} is described in Appendix~\ref{apd:clinician_quality_review}.

\input{tables/mainresults3.tex}

\subsection{Additional validation analyses}
\label{sec:additional_validation_methods}
We perform six additional validation analyses: (1) we quantify uncertainty on \texttt{sepsis-40} using document-level bootstrap resampling; (2) we measure inter-annotator agreement (IAA) on five additional case reports independently annotated by two clinicians; (3) using \texttt{sepsis-10}, we evaluate inference of sepsis onset timestamps from narratives (the onset inference prompt is provided in Appendix~\ref{apd:onset_prompt}); (4) using \texttt{sepsis-10}, we audit extracted events for hallucination by checking whether model-generated findings are grounded in the source report; (5) using the same \texttt{sepsis-10} reports, we evaluate multi-pass self-consistency by re-annotating reports under multiple random seeds; and (6) we estimate full-corpus cross-model agreement between DeepSeek-R1 and Llama-3.3-70B-Instruct across all 2,139 reports. Detailed methods for the hallucination audit, self-consistency, and full-corpus agreement are provided in Appendix~\ref{apd:validation_details}.

\subsubsection{Uncertainty quantification via document-level bootstrap (\texttt{sepsis-40})}
\label{sec:bootstrap_methods}

We use nonparametric bootstrap resampling at the \emph{case report} level with 200 bootstrap replicates.
In each replicate, we sample 40 case reports with replacement from \texttt{sepsis-40} and recompute event match rate, concordance, and AULTC.
We report percentile-based 95\% bootstrap confidence intervals.

\subsubsection{Inter-annotator agreement on additional sepsis reports}
\label{sec:iaa_methods}
Two clinicians independently annotated five sampled T2S2 case reports using the same extraction and timestamping guidelines as \texttt{sepsis-40}.
We compute clinician--clinician agreement and model--clinician agreement using the same metrics (match rate, concordance, and AULTC), enabling comparison between model disagreement and human disagreement under the evaluation protocol.

\subsubsection{Sepsis onset timestamp inference}
\label{sec:onset-inference}
In addition to extracting event--time pairs, we infer an explicit sepsis onset timestamp from narrative cues.
Given a case report, the onset procedure estimates the earliest time consistent with Sepsis-3 criteria relative to admission ($t=0$), producing two onset estimates based on qSOFA and SOFA-style reasoning when sufficient evidence is available.
We used GLM-4.6 (2-bit) for this inference.
The full prompt is provided in Appendix~\ref{apd:onset_prompt}.

\vspace{-1em}
\subsubsection{Manual hallucination audit}
\label{sec:hallucination_audit_methods}
To directly evaluate whether extracted clinical findings were unsupported by the source text, we performed a two-stage hallucination audit on 10 sampled PMOA case reports.
We audited the DeepSeek-R1 and Llama-3.3-70B-Instruct outputs because they represent two strong model families used for corpus-scale annotation and agreement analyses.
First, we applied automated lexical grounding to each extracted event after light normalization, including removal of prompt-induced prefixes such as ``history of'' or ``treated with''.
The grounding step searched for exact substring matches, ordered token subsequences within a sentence, or unordered token containment in the source report.
Second, all events that could not be automatically grounded were manually reviewed against the original case report to determine whether they represented unsupported hallucinations or valid clinical transformations, such as abbreviation expansion, decomposition of conjunctions, or semantic paraphrasing.

\subsubsection{Multi-pass self-consistency}
\label{sec:self_consistency_methods}
To assess robustness across generations, we re-annotated the same 10 reports used in the hallucination audit using five random seeds for DeepSeek-R1, Llama-3.3-70B-Instruct, OpenAI O1, and GPT-5.
For each model, Run 1 was treated as the within-model reference and Runs 2--5 were compared against it using the same event match rate, concordance, and AULTC metrics.
This analysis evaluates whether extracted timelines are stable across stochastic generations rather than being driven by isolated model-specific outputs.

\subsubsection{Full-corpus cross-model agreement}
\label{sec:full_corpus_agreement_methods}
To estimate reliability beyond the human-annotated subset, we measured cross-model agreement between DeepSeek-R1 and Llama-3.3-70B-Instruct across the full 2,139-report T2S2 corpus.
We treated DeepSeek-R1 as the proxy reference and evaluated Llama-3.3-70B-Instruct using event match rate, concordance, and AULTC.
Because these models differ in architecture and training, agreement across the full corpus provides a scalable proxy for whether the extraction rules are applied consistently across independent model families.

\vspace{-1em}

\section{Results}
\vspace{-.5em}
In this section, we first detail the T2S2 corpus and phenotyping accuracy, then evaluate LLM-extracted timelines against clinician annotations and bronze-standard proxy references.
We then present additional validation analyses addressing uncertainty, annotator variability, sepsis onset inference, factual grounding, self-consistency, and full-corpus agreement.

\vspace{-.75em}
\subsection{Corpus overview}
\vspace{-.5em}
Our analysis resulted in T2S2, a new open-access textual time series corpus for Sepsis-3\footnote{Sepsis-3 refers to the Third International Consensus Definitions for Sepsis and Septic Shock \citep{singer2016third}, which define sepsis as a life-threatening organ dysfunction caused by a dysregulated host response to infection.
In contrast, \texttt{sepsis-40}, \texttt{sepsis-10}, and \texttt{sepsis-100}, introduced in this paper, refer to subsets of sepsis reports drawn from PMOA, with the numbers indicating the subset size.}, 
comprising 2,139 case reports.  
Demographically, the corpus includes 58\% male and 42\% female patients, with ages ranging from 0 to 111 and a mean age of 49 (IQR: 32--65).

The \texttt{sepsis-40} subset contains manual clinician annotations for 40 PMOA case reports, yielding a total of $3,229$ manually annotated reference events.
Within this set, \texttt{sepsis-10} denotes the 10-report subset used in our additional validation analysis for clinician annotation quality review, hallucination audit, multi-pass self-consistency, and sepsis onset analysis.
The PMCID composition of \texttt{sepsis-40} and \texttt{sepsis-10} is provided in Appendix~\ref{apd:subset_ids}.
The \texttt{sepsis-100} subset includes 100 T2S2 case reports that do not overlap with \texttt{sepsis-40}, with annotations provided using GPT-5.
We use \texttt{sepsis-100} as a bronze-standard proxy-reference set annotated by GPT-5, not as a clinical gold standard; its purpose is to evaluate extraction behavior at a larger scale than the manually annotated subset while avoiding claims of clinician-level ground truth.

In Appendix~\ref{apd:excerpt}, we provide an expanded multi-model annotation example for PMC10629858 (manual annotations, GPT-5, DeepSeek-R1, Llama 3.3, Mixtral 8x7B, and Llama 3.1 8B).
To complement the aggregate metrics presented in the following sections, Figure~\ref{fig:fine-grained-t2s2-annotation} shows a representative case report with fine-grained timing (PMC3184014) alongside the corresponding T2S2 event--time annotations from Llama 3.3.

\subsection{Phenotyping Results}
For phenotyping the i2m4 dataset for sepsis, we found two case reports of patients with sepsis.
The QwQ-32B-Preview response had perfect accuracy (100\%), whereas the Llama-3.1-8B-Instruct accuracy was 13/20 (65\%).
For the \texttt{sepsis-100} dataset, where QwQ-32B-Preview outputs were used as the preferred computational labels, the relatively faster models GPT-4o and Llama-3.1-8B-Instruct agreed 84\% and 83\% of the time, respectively (Figure~\ref{fig:flow}, bottom right).
Although QwQ-32B-Preview sometimes produced verbose rationales that referenced SOFA-style reasoning, the operational phenotyping decision was mapped to the requested binary output using the rule-based parsing and clinician-audit process described in Section~\ref{section:datasets}.
We therefore interpret these phenotyping labels as cohort-selection outputs for constructing a retrospective research corpus, rather than as a primary clinical phenotyping benchmark.

\subsection{Evaluation of Textual Time Series Corpus for Sepsis}
\subsubsection{Quantitative evaluation of extracted annotations}
\label{sec:annotation_results_quantitative}
Following the quantitative evaluation protocol described in Section~\ref{sec:evaluation_methodology_metrics}, Table~\ref{tab:results} reports performance across i2m4, \texttt{sepsis-40}, and \texttt{sepsis-100}.
On the clinician-annotated \texttt{sepsis-40} set, GPT-5 is the strongest annotator across all four metrics, with event match rate 0.93, concordance 0.965, median absolute error 4 hours, and AULTC 0.822.
DeepSeek-R1 provides the next strongest overall performance on \texttt{sepsis-40}, with higher event recovery and AULTC than Llama 3.3 (event match rate 0.79 vs.\ 0.76; AULTC 0.781 vs.\ 0.742), whereas Llama 3.1 70B and Llama 3.3 show high concordance among their matched events.
This pattern highlights a recurring tradeoff in the evaluation: models can achieve high temporal ordering among the subset of events they match, while still recovering fewer reference events or producing larger absolute timestamp discrepancies.

The i2m4 results should be interpreted as a cross-schema evaluation rather than a direct in-domain test of T2S2-style extraction.
All LLM match rates are lower on i2m4, where the reference annotations are legacy i2b2-style temporal concepts rather than standalone clinical findings.
Among the LLMs, Llama 3.3 performs best on i2m4 across event match rate, concordance, median absolute error, and AULTC, and all LLMs with temporal predictions substantially outperform the BERT 5-CV time-regression baseline on AULTC. However, the low match rates indicate that many clinically meaningful T2S2-style findings are penalized when compared against fragmented concept-level references. 
In contrast to concordances around 0.76 in i2m4, in \texttt{sepsis-40}, GPT-5 and Llama 3.3 were excellent time-ordering annotators, with concordances of 0.965 and 0.908, respectively.

On \texttt{sepsis-100}, which is a bronze-standard proxy-reference set rather than clinical ground truth, DeepSeek-R1 achieves the highest event match rate, lowest median absolute error, and highest AULTC against the GPT-5 reference, while Llama 3.3 achieves the highest concordance.
These results should therefore be read as scalable proxy-reference evaluation of extraction behavior, not as an estimate of absolute clinical accuracy.

To better understand the performance differences across datasets, we conducted an error analysis between \texttt{sepsis-10} and i2m4, presented in Appendix~\ref{apd:error-analysis}.
This analysis shows that the i2m4 performance gap is driven in part by annotation-ontology mismatch between fragmented i2b2 concepts and the more compositional clinical findings targeted by T2S2.

\input{tables/sensitivities3}

\begin{figure*}[tb!]
    \begin{minipage}[t]{0.24\textwidth}
    \centering
    \includegraphics[width=\linewidth, page=1]{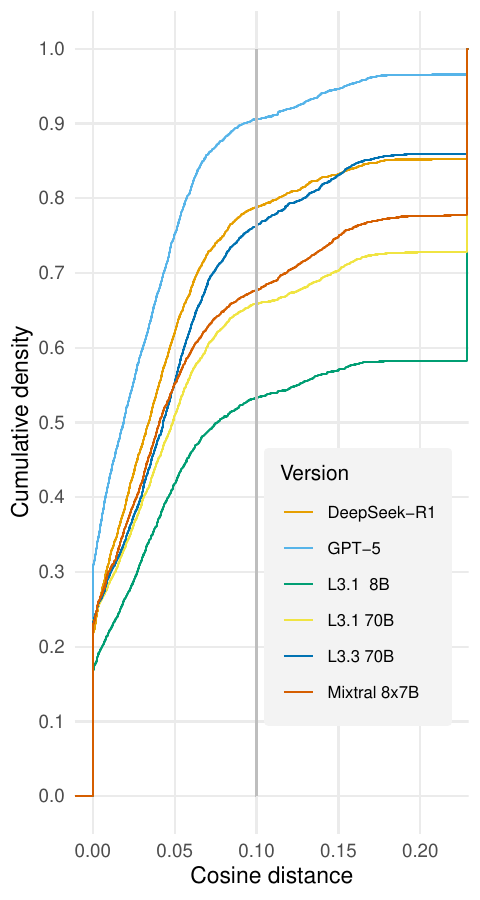}
    \end{minipage}
    \hfill
    \begin{minipage}[t]{0.24\textwidth}
    \centering
    \includegraphics[width=\linewidth, page=2]{figures/main_results_figures/figures_sepsis_40.pdf}
    \end{minipage}
    \begin{minipage}[t]{0.24\textwidth}
    \centering
    \includegraphics[width=\linewidth, page=3]{figures/main_results_figures/figures_sepsis_40.pdf}
    \end{minipage}
    \begin{minipage}[t]{0.24\textwidth}
    \centering
    \includegraphics[width=\linewidth, page=4]{figures/main_results_figures/figures_sepsis_40.pdf}
    \end{minipage}
    \caption{Event match cumulative distribution function (left), concordance box-plots (middle left), 
    time discrepancy from the manual annotation timestamps among matched events, overall (middle right), and disaggregated by clinician annotator timestamp (time from presentation, right) for the \texttt{sepsis-40} dataset.}
    \label{fig:time-disc}
    \vspace{-1em}
\end{figure*}

To examine the results in greater detail, we show figures of the cumulative distribution function of event matches and concordance boxplots (Figure \ref{fig:time-disc}) for the \texttt{sepsis-40} dataset. 
When the function reaches its maximum at the maximum cosine distance, it may be less than 1 because the label events have no events left to match, \emph{i.e.}, the reference method recovered an event that the prediction method did not. 
The event match rates in Table \ref{tab:results} correspond to the match rate at the cosine distance threshold in Figure \ref{fig:time-disc} ($\theta=0.10$).  
One could adjust the cosine distance threshold, for example, by increasing it to increase matches, at the cost of false positive matches that degrade temporal performance. We provide a systematic threshold analysis over $\theta \in [0.01, 0.5]$ in 0.01 increments in Appendix~\ref{apd:threshold_sweep}, which demonstrates a clear elbow near $\theta=0.10$ and supports our chosen operating point.
The concordance box plots (Figure \ref{fig:time-disc}) show strong performance among the large LLMs, whereas the Llama 3.1 8B and Mixtral 8x7B model performances are much lower, which is a consistent finding throughout our analysis.

To better visualize timestamp discrepancies, we plot cumulative time-discrepancy curves among matched events (Figure~\ref{fig:time-disc}).
GPT-5 has the strongest overall time-discrepancy profile, consistent with its median absolute error of 4 hours and highest AULTC in Table~\ref{tab:results}.
The disaggregated curves further show that near-admission events are easier to timestamp than events occurring months or years away from presentation.
This degradation at longer horizons is consistent with the expert qualitative review, which identified ambiguity for events with extended duration and errors in cases requiring chained relative-time reasoning (Appendix~\ref{apd:clinician_quality_review}).
This long-horizon degradation is most visible in the bottom subpanels of Figure~\ref{fig:time-disc} (right), where the strongest models recover a larger fraction of distant events within the correct coarse time scale.

\paragraph{Sensitivity analysis of prompt components.}
The sensitivity analysis to our chosen prompt shows that prompt components affect event recovery and temporal metrics in different ways for Llama-3.3-70B-Instruct (Table~\ref{tab:sens}).
The full prompt remains a strong and balanced operating point, especially for i2m4, where it obtains the highest event match rate, lowest median absolute error, and highest AULTC.
On \texttt{sepsis-40}, removing conjunction expansion preserves the event match rate and improves AULTC, while zero-shot prompting yields higher concordance among matched events but substantially reduces event recovery.
This suggests that high concordance alone can be misleading when fewer or easier-to-order events are matched.
The interval and interval+type augmentations do not consistently improve performance: interval prompting slightly increases event match rate on \texttt{sepsis-100}, but not on \texttt{sepsis-40}, and the additional type field generally does not improve the overall balance of event recovery and temporal accuracy.
Together, these results indicate that adding output fields can increase prompt burden and does not necessarily improve the quality of the extracted textual time series across the models used.
Detailed assessments for the \texttt{sepsis-40} ablations and the i2m4 dataset, analogous to Figure~\ref{fig:time-disc}, are shown in Appendix~\ref{apd:add-perf}.

An expert qualitative review on \texttt{sepsis-10} ranked manual annotations highest, followed by O1-preview, DeepSeek-R1, and Llama 3.3; the full review protocol, ranking table, and representative error modes are provided in Appendix~\ref{apd:clinician_quality_review}.

\subsection{Additional validation and annotation variability}
\label{sec:additional_validation}

\subsubsection{Uncertainty quantification via document-level bootstrap (\texttt{sepsis-40})}
Because clinician-annotated evaluation sets are necessarily limited in size, we quantify uncertainty using nonparametric bootstrap resampling at the \emph{case report} level with 200 bootstrap replicates. We resample case reports with replacement and recompute each metric on each bootstrap replicate, reporting 95\% bootstrap confidence intervals for event match rate, concordance, and AULTC on \texttt{sepsis-40}. Table~\ref{tab:bootstrap_ci} summarizes these intervals. 
Across models, the bootstrap intervals preserve the main qualitative conclusions from Table~\ref{tab:results}: GPT-5 is the strongest model on \texttt{sepsis-40}, DeepSeek-R1 and Llama 3.3 remain the strongest open-model annotators overall, and the smaller or older models generally show lower event recovery and/or temporal accuracy.

\begin{table}[t]
\centering
\small
\caption{Case report bootstrap 95\% confidence intervals (CIs) on \texttt{sepsis-40}. The bootstrap resamples \emph{case reports} with replacement and recomputes metrics per replicate.}
\label{tab:bootstrap_ci}
\adjustbox{max width=\columnwidth}{
\begin{tabular}{lccc}
\toprule
Model & Event match rate & Concordance & AULTC \\
\midrule
GPT-5 & [0.879, 0.930] & [0.956, 0.977] & [0.775, 0.863] \\
DeepSeek-R1 & [0.742, 0.840] & [0.846, 0.929] & [0.724, 0.825] \\
L3.3 70B & [0.731, 0.800] & [0.881, 0.936] & [0.704, 0.790] \\
L3.1 70B & [0.606, 0.705] & [0.897, 0.950] & [0.695, 0.787] \\
Mixtral 8x7B & [0.616, 0.743] & [0.817, 0.935] & [0.667, 0.765] \\
L3.1 8B & [0.483, 0.591] & [0.826, 0.898] & [0.634, 0.736] \\
\bottomrule
\end{tabular}
}
\end{table}

\subsubsection{Inter-annotator agreement on additional sepsis reports}
\label{sec:iaa}
To contextualize residual model--clinician discrepancies, we quantified agreement between two independent clinician annotators on five sampled case reports from T2S2, annotated under the same guidelines. We report (i) model--annotator alignment for two leading LLM annotators against each clinician and (ii) clinician--clinician agreement using the same metrics as the main evaluation.

\begin{table}[!ht]
\centering
\small
\caption{Inter-annotator agreement across reports.}
\label{tab:iaa_manual}
\begin{adjustbox}{width=\columnwidth}
\begin{tabular}{lccc}
\toprule
Annotator pair & Match & Concordance & AULTC \\
\midrule
A1 vs.\ A2 (Manual) & 0.681 & 0.912 & 0.811 \\
\bottomrule
\end{tabular}
\end{adjustbox}
\end{table}

\begin{table}[!ht]
\centering
\small
\caption{Models vs.\ manual annotators (A1, A2).}
\label{tab:iaa_models}
\begin{adjustbox}{width=\columnwidth}
\begin{tabular}{lccc|ccc}
\toprule
& \multicolumn{3}{c|}{Annotator A1} & \multicolumn{3}{c}{Annotator A2} \\
Model & Match & Concordance & AULTC & Match & Concordance & AULTC \\
\midrule
DSR1 & 0.659 & 0.840 & 0.748 & 0.774 & 0.852 & 0.757 \\
L3.3  & 0.594 & 0.875 & 0.784 & 0.723 & 0.872 & 0.759 \\
\bottomrule
\end{tabular}
\end{adjustbox}
\end{table}

Tables~\ref{tab:iaa_manual} and~\ref{tab:iaa_models} contextualize the main results by quantifying annotator variability under the same guidelines and metrics. Clinician--clinician agreement is imperfect even on the same case reports (Table~\ref{tab:iaa_manual}), reflecting inherent ambiguity in mapping narrative descriptions to temporally localized findings and timestamps. Against this reference, the strongest LLM annotators achieve model--clinician agreement that is comparable to clinician--clinician agreement across the same metrics (Table~\ref{tab:iaa_models}), supporting that the leading performance levels in Table~\ref{tab:results} are meaningful relative to human variability rather than artifacts of the evaluation protocol.

\subsubsection{Sepsis onset timestamp inference from narratives}
\label{sec:onset_results}

To broaden T2S2 applicability to onset-anchored analyses, we evaluated an onset inference module that estimates sepsis onset timestamps directly from narrative text.
On \texttt{sepsis-10}, SOFA-based onset matched clinician onset exactly in 6 of 9 valid cases where onset could be assessed, excluding one neonatal outlier for which neonatal SOFA would be more appropriate.
The median absolute error was 0 hours (mean $\approx 3.4$ hours), with a single moderate deviation of 24 hours.
In contrast, qSOFA-based onset tended to lag substantially (mean absolute error $\approx 355$ hours including the outlier and $\approx 126$ hours excluding it), consistent with retrospective reporting of qSOFA-related cues in case narratives rather than true physiologic delay, whereas SOFA-based onset closely aligned with clinician-annotated sepsis onset.

\subsubsection{Additional corpus-scale and \texttt{sepsis-10} validation}
\label{sec:additional_validation_summary}

We further evaluated factual grounding, generation stability, and full-corpus consistency using analyses summarized here and detailed in Appendix~\ref{apd:validation_details}.
In the \texttt{sepsis-10} hallucination audit, 1,517 total extracted events from combined DeepSeek-R1 and Llama-3.3-70B-Instruct were reviewed through automated lexical grounding followed by manual adjudication of unresolved cases; no unsupported hallucinated events were identified (Table~\ref{tab:hallucination_audit} in Appendix~\ref{apd:hallucination_audit_results}).
On the same \texttt{sepsis-10} reports, multi-pass self-consistency across five random seeds showed stable extraction behavior across DeepSeek-R1, Llama 3.3, O1-preview, and GPT-5 (Table~\ref{tab:self_consistency} in Appendix~\ref{apd:self_consistency_results}).
Across the full 2,139-report corpus, Llama-3.3-70B-Instruct achieved strong agreement with DeepSeek-R1, with match rate 0.79, concordance 0.87, and AULTC 0.83 (Table~\ref{tab:full_corpus_agreement} in Appendix~\ref{apd:full_corpus_agreement_results}).

\section{Discussion and Conclusions}
We developed an analytical pipeline for processing free-text biomedical case reports into Sepsis-3 textual time series using large language models. The resulting T2S2 corpus contains 2,139 open-access PMOA case reports represented as event--timestamp trajectories, providing a retrospective resource for temporal modeling of sepsis narratives. Evaluation on the clinician-annotated \texttt{sepsis-40} subset shows strong performance, with GPT-5 achieving the strongest overall results and DeepSeek-R1 and Llama 3.3 providing competitive open-model baselines with different tradeoffs across event recovery, temporal ordering, and absolute timestamp accuracy. 
Additional analyses, including document-level bootstrapping, clinician--clinician agreement, sepsis onset inference, hallucination auditing, self-consistency evaluation, and cross-model agreement on \texttt{sepsis-100} and the full corpus, further contextualize the reliability and limitations of the extracted trajectories.
These results support T2S2 as a retrospective research corpus for downstream tasks such as risk forecasting, causal reasoning, and disease trajectory characterization, while making clear that it is not intended as a deployment-ready clinical extraction system.

There are several limitations to our work.
First, T2S2 is a clinical research informatics corpus constructed from published case reports and should not be considered a representative sample of all patients with sepsis.
Case report corpora are affected by publication bias, rare finding over-representation, selective reporting, and retrospective revisionism.
The corpus is therefore best suited for studying retrospective sepsis trajectories, developing temporal modeling methods, and analyzing narrative representations, rather than estimating population-level incidence or deployment-ready clinical performance.

Second, while \texttt{sepsis-40} provides a substantially expanded human gold standard, exhaustive clinician validation of the full 2,139-report corpus remains infeasible.
The additional validation analyses reduce but do not eliminate this limitation: hallucination auditing provides direct evidence of factual grounding in a sampled subset, self-consistency evaluates stability across repeated generations, and full-corpus cross-model agreement assesses reproducibility across model families.
However, these analyses do not replace clinician adjudication of every extracted event.
A complete characterization of corpus-wide errors would require a larger human-in-the-loop adjudication framework in which clinicians resolve model disagreements and temporal ambiguities.

Third, the substantial use of LLMs in the analytic pipeline can make error analysis difficult because errors may be subtle, clinically plausible, or concentrated in ambiguous temporal expressions.
Our hallucination audit suggests that unsupported event fabrication is uncommon in the audited subset under the extractive formulation, but timestamp errors remain possible even when the event itself is text-grounded.
This distinction is important: the primary failure mode in this task is often not hallucinating a clinical finding, but assigning the wrong time to a real finding, especially when timing requires chained relative reasoning or interpretation of vague duration phrases.

Fourth, the i2m4 analysis should be interpreted as a cross-schema evaluation.
The lower event match rates on i2m4 arise in part from mismatch between fragmented i2b2-style temporal concepts and T2S2-style standalone clinical findings.
For example, section headers, isolated abbreviations, or partial phrases can appear as i2b2 concepts, whereas the T2S2 representation intentionally favors clinically meaningful, compositional findings.
This means that i2m4 is useful as a legacy benchmark and stress test, but it is not a clean estimate of T2S2 extraction performance on clinical notes without re-annotation under the T2S2 schema.

While T2S2 provides a retrospective resource for studying sepsis trajectories, translating this pipeline into a prospective clinical tool would require a different extraction architecture.
Real-time clinical documentation is substantially noisier than finalized case reports or retrospective discharge summaries, and the available information changes continuously as care unfolds.
A prospective system would need to process note updates incrementally, integrate extracted textual findings with streaming structured EHR data such as vitals and laboratory results, calibrate predictions against prospective clinical outcomes, and evaluate safety under deployment-specific error modes.
Until such bridging work is completed, T2S2 should be viewed as a high-fidelity retrospective resource for temporal modeling and clinical trajectory analysis rather than as a clinical decision-support system.

Regarding generalizability to other medical conditions besides sepsis, our methodology could be extended in a straightforward manner. Our motivation for choosing sepsis as the disease focus was to expand the analysis from the i2m4 dataset to a similar population in the PMOA corpus and maintain a well-defined phenotype. Sepsis is a sensible choice as it is one of the most prevalent and important diagnoses in critical care medicine.  
However, there are several concerns when generalizing to other conditions. For one, most other conditions are less acute in nature, meaning that the time distributions may be wider. Also, the amount and type of information in such case reports may vary. 
Despite this, one benefit of AULTC compared to linear or squared discrepancies is that when intensive-care and chronic-disease cases are mixed, long-horizon discrepancies from chronic disease cases do not dominate the performance measure because of the log-time scaling.
Another generalizability concern is that for some conditions, less temporal information will be present. This appears to be more acute in imaging studies, where the focus of the case is on the image content at a single snapshot in time. For these types of studies, the longitudinal annotations become less meaningful compared to the image's information content.

Future work could focus on expanding this methodological framework. While our current prompt augmentations did not yield systematic performance gains (Table \ref{tab:sens}), exploring advanced prompting paradigms alongside domain-specific fine-tuning could further refine the fidelity and expressivity of the extracted representations. We anticipate that T2S2 will serve as a foundational resource for downstream healthcare applications, including causal inference, risk forecasting, and multimodal alignment. The empirical utility of this representation is evidenced by our concurrent work, which leverages T2S2 from clinical narratives to accurately forecast future patient risk states \citep{noroozizadeh2026forecasting}. Ultimately, advancing temporal AI in healthcare requires exploiting all available modalities; the extraction of structured temporal trajectories from narrative text provides a critical complement to existing structured data streams.

Subsequent iterations of this pipeline can be tailored to specific downstream objectives, each imposing distinct constraints. For instance, causal modeling requires strict time-ordering, whereas forecasting demands precise absolute timestamps. Furthermore, while our current formulation targets true physiological occurrence, an alternative paradigm could focus on the timing of clinical recognition. Capturing these perceived occurrences could illuminate decision-making dynamics and facilitate research in dynamic resource allocation. 
Ultimately, integrating these narrative-derived timelines with structured physiological data offers a promising avenue to build more comprehensive, multimodal representations of patient trajectories.

\acks{
This research was supported in part by the Division of Intramural Research (DIR) of the National Library of Medicine (NLM), National Institutes of Health.
This work utilized the computational resources of the NIH HPC Biowulf cluster.
S.N. was supported by Carnegie Mellon University TCS Presidential Fellowship, and Natural Sciences and Engineering Research Council of Canada (NSERC) Canada Graduate Research Scholarship --- Doctoral (CGRS D). S.N. was also supported in part by an appointment to the National Library of Medicine Research Participation Program administered by the Oak Ridge Institute for Science and Education (ORISE) through an interagency agreement between the U.S. Department of Energy (DOE) and the National Library of Medicine, National Institutes of Health. ORISE is managed by ORAU under DOE contract number DE-SC0014664. All opinions expressed in this paper are the authors’ and do not necessarily reflect the policies and views of NIH, NLM, DOE, or ORAU/ORISE.}

\bibliography{refs}

\clearpage
\numberwithin{equation}{section}
\numberwithin{figure}{section}
\numberwithin{table}{section}
\numberwithin{algorithm}{section}
\appendix

\onecolumn        
\section*{Appendix}
\addcontentsline{toc}{section}{Appendix Overview}

This appendix is organized as follows:

\begin{itemize}
    \item \textbf{Appendix~\ref{apd:subset_ids} (\nameref{apd:subset_ids}):} PMCID lists for the \texttt{sepsis-40} and \texttt{sepsis-10} annotated subsets.

    \item \textbf{Appendix~\ref{apd:clinician_quality_review} (\nameref{apd:clinician_quality_review}):} Expert qualitative review protocol, ranking results, and representative annotation error modes for \texttt{sepsis-10}.

    \item \textbf{Appendix~\ref{apd:prompt} (\nameref{apd:prompt}):} Main clinical event--timestamp extraction prompt.

    \item \textbf{Appendix~\ref{apd:phenotype-prompt} (\nameref{apd:phenotype-prompt}):} Sepsis-3 phenotyping prompt used for computational labeling.

    \item \textbf{Appendix~\ref{apd:onset_prompt} (\nameref{apd:onset_prompt}):} Prompt for sepsis onset timestamp inference.

    \item \textbf{Appendix~\ref{apd:ltcdf} (\nameref{apd:ltcdf}):} Derivation and discussion of the area under the log-time curve (AULTC) metric.

    \item \textbf{Appendix~\ref{apd:pseudo} (\nameref{apd:pseudo}):} Pseudocode for the recursive best-match procedure used for event alignment.

    \item \textbf{Appendix~\ref{apd:excerpt} (\nameref{apd:excerpt}):} Worked examples with manual and LLM annotations, including an illustrative example with finer-grained timing (Appendix~\ref{apd:fine_example}).

    \item \textbf{Appendix~\ref{apd:threshold_sweep} (\nameref{apd:threshold_sweep}):} Sensitivity analysis of the cosine distance matching threshold used to compute event match rate.

    \item \textbf{Appendix~\ref{apd:add-perf} (\nameref{apd:add-perf}):} Additional performance analyses and ablation studies.

    \item \textbf{Appendix~\ref{apd:validation_details} (\nameref{apd:validation_details}):} Additional details for the hallucination audit, multi-pass self-consistency analysis, and full-corpus cross-model agreement.

    \item \textbf{Appendix~\ref{apd:error-analysis} (\nameref{apd:error-analysis}):} Error analysis of event matching and temporal predictions for i2m4 versus \texttt{sepsis-40} performance differences.
\end{itemize}

\clearpage

\section{Annotated PMOA Subsets}
\label{apd:subset_ids}

The \texttt{sepsis-40} subset contains 40 PMOA case reports with manual clinician annotations, yielding $3,229$ reference clinical events.
The \texttt{sepsis-10} subset is a 10-report subset (801 reference events) of \texttt{sepsis-40} used for clinician annotation quality review, hallucination audit, multi-pass self-consistency, and sepsis onset analysis.

\begin{table*}[h]
\centering
\caption{\hspace*{0.08\linewidth}\begin{minipage}{0.85\linewidth}
PMCID composition of the annotated PMOA subsets for \texttt{sepsis-40}.\\
A checkmark indicates membership in \texttt{sepsis-10}.
\end{minipage}}
\label{tab:subset_pmcs}
\begin{tabular}{lc@{\hspace{3em}}lc}
\toprule
PMCID & \texttt{sepsis-10} & PMCID & \texttt{sepsis-10} \\
\midrule
PMC10037274 &  & PMC10493687 &  \\
PMC3075162  & \checkmark & PMC4421677  &  \\
PMC4929359  &  & PMC5862012  &  \\
PMC6345085  &  & PMC7331532  &  \\
PMC8546416  &  & PMC9355071  & \checkmark \\
PMC10370906 & \checkmark & PMC10556716 & \checkmark \\
PMC3130983  &  & PMC4573216  &  \\
PMC5142474  &  & PMC6059163  &  \\
PMC6441647  &  & PMC7451141  &  \\
PMC8855197  &  & PMC9393191  &  \\
PMC10406025 &  & PMC10629858 & \checkmark \\
PMC3325348  &  & PMC4778217  & \checkmark \\
PMC5721301  &  & PMC6231299  &  \\
PMC6910033  &  & PMC7576661  & \checkmark \\
PMC9136437  & \checkmark & PMC9552858  & \checkmark \\
PMC10480000 &  & PMC2698425  &  \\
PMC4120057  &  & PMC4898175  &  \\
PMC5747955  &  & PMC6238558  & \checkmark \\
PMC7093739  &  & PMC8298783  &  \\
PMC9260376  &  & PMC9647393  &  \\
\bottomrule
\end{tabular}
\end{table*}

\clearpage

\section{Clinician Annotation Quality Review}
\label{apd:clinician_quality_review}

\subsection{Review protocol}

In addition to the quantitative metrics used in the main evaluation, a physician reviewed extracted timelines for the \texttt{sepsis-10} subset.
For each case report, the clinician assigned: (i) a rank ordering (1 = best, 7 = worst) of the annotations, consisting of the manual annotation and six LLM-generated annotations, and (ii) a categorical assessment (Excellent, Good, Acceptable, or Poor) based on extraction completeness, temporal ordering accuracy, and timing plausibility.
This review provides a qualitative assessment of annotation style and clinical usability, complementing the quantitative event-matching and temporal-alignment metrics reported in the main text.
The reviewed annotation bundle included O1-preview as the OpenAI annotation; replacing GPT-5 which was evaluated separately in the main quantitative results in Table~\ref{tab:results}.

\subsection{Clinician review results}

Inspecting the clinical quality of the annotations, we report the mean rank, top-1 fraction, top-1-LLM fraction, and quality level in Table~\ref{tab:sep10review}.
There is a clear performance ordering in the review, with manual annotations ranked highest, O1-preview a close second, followed by DeepSeek-R1 IQ1 and Llama 3.3.
While the Llama 3.3 annotations were rated acceptable a high percentage of the time (90\%), the other Llama and Mixtral approaches fail to achieve Acceptable performance in excess of 50\%.
Comparing DeepSeek-R1 IQ1 and Llama 3.3, the former has more variable clinician-rated quality (8 of 10 cases rated Acceptable or better), but maintains both higher match rate and AULTC in the main quantitative evaluation (Table~\ref{tab:results} and Figure~\ref{fig:time-disc}).
The clinical reviewer also noted that some Llama 3.3 discrepancies for large time horizons, \emph{i.e.}, years, were large, which is corroborated by the long-horizon subpanels in Figure~\ref{fig:time-disc}.

\begin{table}[h]
\centering
\small
\caption{Clinician annotation quality review on \texttt{sepsis-10}.}
\label{tab:sep10review}
\begin{tabular}{lcccccc}
\toprule
Annotator & Mean Rank & Top-1 & Top-1-LLM & Excellent & $\geq$ Good & $\geq$ Acceptable \\
\midrule
Manual & 1.2 & 0.9 & -- & 1 & 1 & 1 \\
O1-preview & 1.5 & 0.6 & 0.9 & 0.9 & 1 & 1 \\
DeepSeek R1 IQ1 & 3.5 & 0.1 & 0.2 & 0.4 & 0.5 & 0.8 \\
L3.3 70B & 3.8 & 0 & 0 & 0 & 0.2 & 0.9 \\
L3.1 70B & 4.8 & 0 & 0 & 0 & 0.2 & 0.5 \\
Mixtral 8x7B & 5.4 & 0 & 0 & 0 & 0.1 & 0.2 \\
L3.1 8B & 6.5 & 0 & 0 & 0 & 0 & 0 \\
\bottomrule
\end{tabular}
\end{table}

The annotation review feedback noted a wide variety of errors; here we present several error modes identified during quality review.

\paragraph{Events with duration.}
Example: ``10-kg weight loss'' from ``he was presented to the hospital with complaints of abdominal distension, constipation, vomiting, and a 10-kg weight loss.''
This example from case report PMC10629858 references a gradual event that has been ongoing for 2 months.
By the annotation guideline for selecting the start time of events with duration, the time -168 (what O1-preview annotates) is correct.
However, the event can also be considered as ``history of 10-kg weight loss'' via the ``history of'' guideline with time 0 (what L3.3 70B annotates).
This example could be resolved by interval representations, but we see the annotation quality, particularly AULTC, degrade when requesting additional fields (Table~\ref{tab:sens} and Figure~\ref{fig:sep40ablate-time-disc}).

\paragraph{Events requiring temporal reasoning or composition.}
Example: ``passed away'' from ``he was re-admitted to the medical ICU for severe sepsis and multiorgan failure and passed away around 6 months after his initial diagnosis with NHL, despite maintaining a remission status.''
This requires a two-step inference: (1) timing of NHL diagnosis, and (2) adding ``around 6 months'' to that time.
While this is a simple example, more complicated chains of relative timing were noted, sometimes involving other events whose timings are themselves uncertain.
As black-box methods, there is no explicit probabilistic specification for these chains, and empirically we see variable responses.
For this case, the manual annotation is 4383 hours (about 6 months), O1-preview gives 4368 hours (about 6 months), DeepSeek-R1 gives 4032 hours (about 6 months), and Llama 3.3 70B gives 1440 hours (about 2 months); compared to other events the temporal ordering is correct, but the relative timing is off by months.

\clearpage

\section{Annotation Prompt}\label{apd:prompt}
The main annotation prompt used for clinical finding extraction is shown below. 
Specific components are color-coded to relate them directly to the ablation analysis in Section~\ref{sec:annotation_results_quantitative} and Table~\ref{tab:sens}: \textcolor{darkpurple}{role-playing instruction} (``You are a physician''), \textcolor{darkred}{conjunction expansion}, and \textcolor{darkteal}{few-shot example}.

\begin{tcolorbox}[
  colback=teal!5,
  colframe=teal!70!black,
  title=Textual Time Series Generation Prompt,
  boxrule=2pt,
]
\footnotesize
\setlength{\parskip}{1pt}
\color{darkpurple} You are a physician. 

\color{black}  Extract the clinical events and the related time stamp from the case report. The admission event has timestamp 0. If the event is not available, we treat the event, e.g. current main clinical diagnosis or treatment with timestamp 0. The events happened before event with 0 timestamp have negative time, the ones after the event with 0 timestamp have positive time. The timestamp are in hours. The unit will be omitted when output the result. If there is no temporal information of the event, please use your knowledge and events with temporal expression before and after the events to provide an approximation. We want to predict the future events given the events happened in history. 

\color{darkteal} For example, here is the case report.

An 18-year-old male was admitted to the hospital with a 3-day history of fever and rash. Four weeks ago, he was diagnosed with acne and received the treatment with minocycline, 100 mg daily, for 3 weeks. With increased WBC count, eosinophilia, and systemic involvement, this patient was diagnosed with DRESS syndrome. The fever and rash persisted through admission, and diffuse erythematous or maculopapular eruption with pruritus was present. One day later the patient was discharged.

Let's find the locations of event in the case report, it shows that four weeks ago of fever and rash, four weeks ago, he was diagnosed with acne and receive treatment. So the event of fever and rash happen four weeks ago, 672 hours, it is before admitted to the hospital, so the time stamp is -672. diffuse erythematous or maculopapular eruption with pruritus was documented on the admission exam, so the time stamp is 0 hours, since it happens right at admission. DRESS syndrome has no specific time, but it should happen soon after admission to the hospital, so we use our clinical judgment to give the diagnosis of DRESS syndrome the timestamp 0. Then the output should look like

18 years old $|$ 0\

male $|$ 0\

admitted to the hospital $|$ 0\

fever $|$ -72\

rash $|$ -72\

acne $|$  -672\

minocycline $|$  -672\

increased WBC count $|$ 0\

eosinophilia$|$ 0\

systemic involvement$|$ 0\

diffuse erythematous or maculopapular eruption$|$ 0\

pruritis $|$ 0\

DRESS syndrome $|$ 0\

fever persisted $|$ 0\

rash persisted $|$ 0\

discharged $|$ 24\ 

\color{darkred}
Separate conjunctive phrases into its component events and assign them the same timestamp (for example, the separation of `fever and rash' into 2 events: `fever' and `rash'). 

\color{black} If the event has duration, assign the event time as the start of the time interval. Attempt to use the text span without modifications except `history of' where applicable. Include all patient events, even if they appear in the discussion; do not omit any events; include termination/discontinuation events; include the pertinent negative findings, like `no shortness of breath' and `denies chest pain'.  Show the events and timestamps in rows, each row has two columns: one column for the event, the other column for the timestamp.  The time is a numeric value in hour unit. The two columns are separated by a pipe `$|$' as a bar-separated file. Skip the title of the table. Reply with the table only. Create a table from the following case: 

\textless \textbackslash end prefix\textgreater

\end{tcolorbox}

\clearpage

\section{Phenotyping Prompt}\label{apd:phenotype-prompt}
For Sepsis-3 determination, we had a clinician review the discharge summaries and assign each case as meeting the Sepsis-3 definition.  
To develop a computational approach, we constructed a prompt specific to Sepsis-3, which describes the qSOFA procedure for determining a Sepsis-3 diagnosis, and used it as query to Llama-3.1-8B-Instruct (L3.1 8B) and QwQ-32B-Preview (Qwen).  
We used the phenotyping prompt:

\begin{tcolorbox}[
  colback=yellow!10,
  colframe=orange!80!black,
  title=Phenotyping Prompt,
  boxrule=2pt
]
    {You are an expert physician.}  
    
    {Determine if the patient described in the following case report has either sepsis or septic shock, as defined by the Sepsis-3 criteria, which correspond to having a} 
    
    {(1) suspected or confirmed infection and} 
    
    {(2) blood pressure/respiratory rate/mental status abnormalities.} 
    
    {If the information is not present, use your best judgment based on the information available.} 
    
    {Reply 1 for sepsis, 0 otherwise.} 
    
    {Reply with the number 0 or 1 only in \textbackslash\!boxed\{\textbackslash\!n \boxed{<0 $ or $ 1>} \textbackslash\!n\}with no explanation.}
    
    {Here is the case:} 
    
    {\textless \textbackslash end prefix\textgreater}
\end{tcolorbox}

\clearpage

\section{Sepsis onset inference prompt}
\label{apd:onset_prompt}
We estimate sepsis onset time relative to admission ($t=0$) using a prompt that requests two onset times: one using qSOFA and one using SOFA-style reasoning when sufficient evidence is available. The model outputs a two-row bar-separated table with onset timestamps in hours (or NA if criteria are not met).

\begin{tcolorbox}[
  colback=yellow!10,
  colframe=orange!80!black,
  title=Sepsis Onset Prompt,
  boxrule=2pt
]
    You are an astute physician. 
    
    Determine the time of onset of sepsis using two distinct definitions:
    
    (1) using qSOFA, and 
    
    (2) using SOFA, 
    
    relative to the time of admission which has timestamp 0. 
    
    If the admission time is not available, treat the event, e.g. current main clinical diagnosis or treatment with timestamp 0. The events happened before event with 0 timestamp have negative time, the ones after the event with 0 timestamp have positive time. The timestamps are in hours. The unit will be omitted when output the result. 
    
    If there is no temporal information regarding the onset times, please use your knowledge and events with temporal expression before and after the events to provide an approximation. 
    
    Similarly if the numeric quantities in qSOFA/SOFA are not present, use the available text to infer them as needed.
    
    Show the sepsis onsets and timestamps in rows, each row has two columns: one column for the sepsis definition used (qSOFA or SOFA), the other column for the timestamp. 
    
    The time is a numeric value in hour unit. If criteria is not met, set the time to NA. The two columns are separated by a pipe $|$ as a bar-separated file. Skip the title of the table. Create the table from the following case:

    {\textless \textbackslash end prefix\textgreater}

\end{tcolorbox}

\include{sections/aultc}

\section{Recursive Best Match Procedure}
\label{apd:pseudo}
We provide pseudocode for the best match procedure between two lists of strings (Algorithm \ref{alg:recursive_match}).  For each list, we use the text order, \emph{i.e.}, the order of the events in the annotation files, to break embedding distance ties.  We use the cosine distance for the distance calculation using sentence transformer embeddings from \texttt{S-PubMedBert-MS-MARCO}.

\begin{algorithm*}[!ht]
\small
\caption{Recursive Best Match}
\label{alg:recursive_match}
\SetAlgoLined
\SetKwFunction{FnMatchEvents}{MatchEvents}
\SetKwInOut{Input}{Input}
\SetKwInOut{Output}{Output}

\Input{\quad Two lists: \texttt{ref} (reference events) and \texttt{pred} (predicted events)}
\Output{\quad List of best-matching event pairs}
\FnMatchEvents{\texttt{ref}, \texttt{pred}} {
    \; 

    \Indp
    \If{ref is empty \textbf{or} pred is empty}{
        \Return{\textbf{[]}}
    }
    Initialize $\text{min\_distance} \gets \infty$\;
    
    Initialize $\text{best\_pair} \gets \text{None}$\;
    
    \ForEach{$r$ \textbf{in} ref}{
        \ForEach{$p$ \textbf{in} pred}{
            $d \gets \text{ComputeDistance}(r, p)$\;
            
            \If{$d < \text{min\_distance}$}{
                $\text{min\_distance} \gets d$\;
                
                $\text{best\_pair} \gets (r, p)$\;
            }\ElseIf{$d = \text{min\_distance}$}{
                $\text{current\_ref\_index} \gets \text{index of } r \text{ in ref}$\;
                
                $\text{current\_pred\_index} \gets \text{index of } p \text{ in pred}$\;
                
                $\text{best\_ref\_index} \gets \text{index of best\_pair.r in ref}$\;
                
                $\text{best\_pred\_index} \gets \text{index of best\_pair.p in pred}$\;
                
                \If{$\text{current\_ref\_index} < \text{best\_ref\_index}$}{
                    $\text{best\_pair} \gets (r, p)$\;
                }\ElseIf{$\text{current\_ref\_index} = \text{best\_ref\_index}$ \textbf{and} $\text{current\_pred\_index} < \text{best\_pred\_index}$}{
                    $\text{best\_pair} \gets (r, p)$\;
                }
            }
        }
    }
    Remove $\text{best\_pair.r}$ from ref\;
    
    Remove $\text{best\_pair.p}$ from pred\;
    
    $\text{result} \gets [\text{best\_pair}] + \text{MatchEvents}
    (\text{ref}, \text{pred})$\;
    
    \Return{$\text{result}$}\;
}
\end{algorithm*}

\clearpage

\section{Example Annotations}\label{apd:excerpt}
In this section we present an excerpt from PMC10629858, the most recently published case in the \texttt{sepsis-40} dataset \citep{abu2023rare}. To illustrate the extraction task and performance characteristics of the LLM annotators, we contrast the manual annotations with that of GPT-5, Llama 3.3, and Llama 3.1.

\begin{figure*}[!h]
\centering

\begin{tcolorbox}[
  enhanced,
  colback=gray!4,
  colframe=gray!35,
  boxrule=0.4pt,
  arc=2mm,
  left=3mm,right=3mm,top=2mm,bottom=2mm,
]

{\bfseries Excerpt from \cite{abu2023rare}:}
\vspace{2pt}

\begin{tcolorbox}[
  enhanced,
  colback=blue!3,
  colframe=blue!60!black,
  title=\textbf{PMC10629858},
  fonttitle=\small,
  boxrule=0.8pt,
  arc=2mm,
  left=3mm,right=3mm,top=2mm,bottom=2mm
]

\footnotesize

A 57-year-old man recently diagnosed with lepromatous leprosy was confirmed with skin biopsy and had been on treatment (rifampicin/clofazimine/dapsone) for 2 months before admission; he was presented to the hospital with complaints of abdominal distension, constipation, vomiting, and a 10-kg weight loss. On examination, the patient was vitally stable. He had evidence of peripheral lymphadenopathy with a distended abdomen and a positive shifting dullness. A computed tomography scan of his abdomen showed mural thickening of the terminal ileum with significantly enlarged mesenteric lymph nodes, mesenteric fat stranding, and intra-abdominal free fluid, suggesting abdominal granulomatous infection or neoplastic process.

\dots

The patient was planned for consolidation by autologous bone marrow transplant. Unfortunately, with the recurrent bacteremia and sepsis that accompanied the patient’s course due to his immunocompromised state, he was re-admitted to the medical ICU for severe sepsis and multiorgan failure and passed away around 6 months after his initial diagnosis with NHL, despite maintaining a remission status.

\end{tcolorbox}

\vspace{3mm}

\begin{minipage}[t]{0.49\textwidth}
\vspace{0pt} %
\begin{tcolorbox}[annobox,title={Manual annotation}]
\annottablestyle
\begin{tabular}{P{0.78\textwidth} r}
\toprule
\textbf{Event} & \textbf{Time} \\
\midrule
57-year-old&0 \\
man&0 \\
diagnosed with lepromatous leprosy&-1461 \\
on treatment&-1461 \\
rifampicin&-1461 \\
clofazimine&-1461 \\
dapsone&-1461 \\
presented to the hospital&0 \\
abdominal distension&0 \\
constipation&0 \\
vomiting&0 \\
10-kg weight loss&0 \\
vitally stable&0 \\
peripheral lymphadenopathy&0 \\
distended abdomen&0 \\
positive shifting dullness&0 \\
computed tomography scan of his abdomen&0 \\
mural thickening of the terminal ileum&0 \\
significantly enlarged mesenteric lymph nodes&0 \\
mesenteric fat stranding&0 \\
intra-abdominal free fluid&0 \\
re-admitted to the medical ICU&4383 \\
severe sepsis&4383 \\
multiorgan failure&4383 \\
passed away&4383 \\
\bottomrule
\end{tabular}
\end{tcolorbox}
\end{minipage}
\hfill
\begin{minipage}[t]{0.49\textwidth}
\vspace{0pt}
\begin{tcolorbox}[annobox,title={GPT-5 annotation}]
\annottablestyle
\begin{tabular}{P{0.78\textwidth} r}
\toprule
\textbf{Event} & \textbf{Time} \\
\midrule
57-year-old man & 0 \\
Male & 0 \\
Lepromatous leprosy & -1440 \\
Skin biopsy confirmed lepromatous leprosy & -1440 \\
Started on rifampicin & -1440 \\
Started on clofazimine & -1440 \\
Started on dapsone & -1440 \\
10-kg weight loss & -720 \\
Presented to the hospital & 0 \\
Complaints of abdominal distension & 0 \\
Complaints of constipation & 0 \\
Complaints of vomiting & 0 \\
Patient was vitally stable & 0 \\
Peripheral lymphadenopathy & 0 \\
Distended abdomen & 0 \\
Positive shifting dullness & 0 \\
Computed tomography scan of abdomen & 6 \\
Mural thickening of the terminal ileum & 6 \\
Significantly enlarged mesenteric lymph nodes & 6 \\
Mesenteric fat stranding & 6 \\
Intra-abdominal free fluid & 6 \\
...\\
Bone marrow examination performed & 96 \\
...\\
Re-admitted to the medical ICU & 4380 \\
Severe sepsis & 4380 \\
Multiorgan failure & 4380 \\
Maintaining remission status & 4380 \\
Passed away & 4392 \\
\bottomrule
\end{tabular}
\end{tcolorbox}
\end{minipage}

\end{tcolorbox}
\end{figure*}

\begin{figure*}[!ht]
\centering
\begin{adjustbox}{max width=\linewidth, max totalheight=0.5\textheight}
\begin{tcolorbox}[annopanel]

\begin{minipage}[t]{0.49\textwidth}
\vspace{0pt}   %
\begin{tcolorbox}[annobox,title={DeepSeek-R1-IQ1 annotation}]
\annottablestyle
\begin{tabular}{P{0.78\textwidth} r}
\toprule
\textbf{Event} & \textbf{Time} \\
\midrule
57 years old & 0 \\
male & 0 \\
lepromatous leprosy & -1464 \\
skin biopsy & -1464 \\
rifampicin & -1464 \\
clofazimine & -1464 \\
dapsone & -1464 \\
admitted to the hospital & 0 \\
abdominal distension & 0 \\
constipation & 0 \\
vomiting & 0 \\
10-kg weight loss & 0 \\
vital stability & 0 \\
peripheral lymphadenopathy & 0 \\
distended abdomen & 0 \\
positive shifting dullness & 0 \\
computed tomography scan & 0 \\
mural thickening of the terminal ileum & 0 \\
enlarged mesenteric lymph nodes & 0 \\
mesenteric fat stranding & 0 \\
intra-abdominal free fluid & 0 \\
abdominal granulomatous infection or neoplastic process & 0 \\
plan for autologous bone marrow transplant & 0 \\
recurrent bacteremia & 0 \\
sepsis & 0 \\
readmission to medical ICU & 0 \\
severe sepsis & 24 \\
multiorgan failure & 24 \\
passed away & 4032 \\
\bottomrule
\end{tabular}
\end{tcolorbox}
\end{minipage}
\hfill
\begin{minipage}[t]{0.49\textwidth}
\vspace{0pt}   %
\begin{tcolorbox}[annobox,title={Llama-3.3-70B-Instruct annotation}]
\annottablestyle
\begin{tabular}{P{0.78\textwidth} r}
\toprule
\textbf{Event} & \textbf{Time} \\
\midrule
57 years old & 0 \\
male & 0 \\
lepromatous leprosy & -720 \\
treatment with rifampicin/clofazimine/dapsone & -720 \\
abdominal distension & 0 \\
constipation & 0 \\
vomiting & 0 \\
10-kg weight loss & 0 \\
peripheral lymphadenopathy & 0 \\
distended abdomen & 0 \\
positive shifting dullness & 0 \\
mural thickening of the terminal ileum & 0 \\
enlarged mesenteric lymph nodes & 0 \\
mesenteric fat stranding & 0 \\
intra-abdominal free fluid & 0 \\
abdominal granulomatous infection & 0 \\
neoplastic process & 0 \\
consolidation by autologous bone marrow transplant & 720 \\
recurrent bacteremia and sepsis & 720 \\
multiorgan failure & 720 \\
death & 1440 \\
\bottomrule
\end{tabular}
\end{tcolorbox}
\end{minipage}

\end{tcolorbox}
\end{adjustbox}
\end{figure*}

\begin{figure*}[!ht]
\centering
\begin{adjustbox}{max width=\linewidth, max totalheight=0.45\textheight}
\begin{tcolorbox}[annopanel]

\begin{minipage}[t]{0.49\textwidth}
\vspace{0pt}   %
\begin{tcolorbox}[annobox,title={Mixtral-8x7B-Instruct v0.1 annotation}]
\annottablestyle
\begin{tabular}{P{0.78\textwidth} r}
\toprule
\textbf{Event} & \textbf{Time} \\
\midrule
57 years old& 0 \\
male & 0 \\
recently diagnosed with lepromatous leprosy & -60 \\
on treatment (rifampicin/clofazimine/dapsone) & -60 \\
presented to the hospital & 0 \\
complaints of abdominal distension & 0 \\
complaints of constipation & 0 \\
complaints of vomiting & 0 \\
complaints of 10-kg weight loss & 0 \\
evidence of peripheral lymphadenopathy & 0 \\
distended abdomen & 0 \\
positive shifting dullness & 0 \\
CT showed mural thickening of the terminal ileum & 0 \\
CT showed significantly enlarged mesenteric lymph nodes & 0 \\
CT showed mesenteric fat stranding & 0 \\
CT showed intra-abdominal free fluid & 0 \\
showing a dramatic response to treatment & 0 \\
planned for consolidation by autologous bone marrow transplant & 0 \\
re-admitted to the medical ICU for severe sepsis and multiorgan failure & 0 \\
passed away around 6 months after his initial diagnosis with NHL & 0 \\
despite maintaining a remission status & 0 \\
\bottomrule
\end{tabular}
\end{tcolorbox}
\end{minipage}
\hfill
\begin{minipage}[t]{0.49\textwidth}
\vspace{0pt}   %
\begin{tcolorbox}[annobox,title={Llama-3.1-8B-Instruct annotation}]
\annottablestyle
\begin{tabular}{P{0.78\textwidth} r}
\toprule
\textbf{Event} & \textbf{Time} \\
\midrule
18 years old & 0 \\
male & 0 \\
admitted to the hospital & 0 \\
lepromatous leprosy & -720 \\
treatment with rifampicin/clofazimine/dapsone & -720 \\
abdominal distension & -720 \\
constipation & -720 \\
vomiting & -720 \\
weight loss & -720 \\
peripheral lymphadenopathy & -720 \\
distended abdomen & -720 \\
positive shifting dullness & -720 \\
mural thickening of the terminal ileum & -720 \\
enlarged mesenteric lymph nodes & -720 \\
mesenteric fat stranding & -720 \\
intra-abdominal free fluid & -720 \\
abdominal granulomatous infection or neoplastic process & -720 \\
autologous bone marrow transplant & -720 \\
recurrent bacteremia & -720 \\
sepsis & -720 \\
multiorgan failure & -720 \\
passed away & -720 \\
\bottomrule
\end{tabular}
\end{tcolorbox}
\end{minipage}

\end{tcolorbox}
\end{adjustbox}
\end{figure*}

\clearpage
\subsection{Additional illustrative example with finer-grained timing}

In this section, we present an additional example representative of case reports with finer-grained temporal detail. We also show the T2S2 annotation produced by Llama 3.3 alongside the case report to illustrate the extracted timeline for this case report.

\label{apd:fine_example}
\begin{figure*}[!ht]
\centering

\begin{tcolorbox}[annopanel]

\begin{minipage}[t]{0.48\textwidth}
\vspace{0pt} %
\begin{tcolorbox}[
  enhanced,
  colback=blue!3,
  colframe=blue!60!black,
  title=\textbf{PMC3184014},
  fonttitle=\small,
  boxrule=0.8pt,
  arc=2mm,
  left=3mm,right=3mm,top=2mm,bottom=2mm
]
\footnotesize
A 39-year-old woman with a history of type 2 diabetes, deep vein thrombosis, gastritis, schizophrenia and self-harm presented to the medical admissions unit with a one-week history of a right sub-mammary abscess followed by spreading cellulitis of the breast. On examination she was septic with a pyrexia of 39.2, blood pressure of 127/65 and tachycardia of 110. Examination of the breast revealed widespread cellulitis involving the nipple and sub-mammary area spreading to the axilla. Initial blood tests showed a white cell count of 23.8 (neutrophils 19.9) and CRP of 428. Simple cellulitis was diagnosed by the admitting physicians and she was commenced on parenteral antibiotics and fluid resuscitation. Her condition worsened despite intravenous benzyl penicillin and flucloxacillin. A surgical opinion was sought on day three when she had evidence of synergistic gangrene. The cellulitis had now spread and there were areas of growing necrotic ulceration (Figure~1). Resuscitation was commenced and antibiotic therapy was adjusted to imipenem and clindamycin based on local necrotizing fasciitis guidelines. She was taken to theatre within a few hours where a partial mastectomy was performed and the wound was left open and packed (Figure~2). Postoperatively she remained stable on intensive care. The following day she returned to theatre for further debridement.

The wound was monitored closely and a vacuum dressing was applied to aid healing. Signs of sepsis improved and she returned to the ward on the fifth postoperative day. Nutritional supplements and adequate hydration were continued during her recovery. Secondary closure was performed on day 13 after the initial operation and there was no further breakdown or compromise of the remaining tissues.

Preoperative cultures taken from the ulcers identified a mixture of gram positive and negative bacteria including \textit{Bacteroides} spp. Histology confirmed widespread microscopic changes and abscess formation consistent with gangrene. There was no evidence of malignancy. She was discharged 22 days after admission and follow-up was arranged.
\end{tcolorbox}
\end{minipage}
\hfill
\begin{minipage}[t]{0.48\textwidth}
\vspace{0pt} %
\begin{tcolorbox}[annobox,title={T2S2 annotation}]
\annottablestyle
\begin{tabular}{P{0.78\textwidth} r}
\toprule
\textbf{Event} & \textbf{Time} \\
\midrule
39 years old & 0 \\
woman & 0 \\
type 2 diabetes & -168 \\
deep vein thrombosis & -168 \\
gastritis & -168 \\
schizophrenia & -168 \\
self-harm & -168 \\
presented to the medical admissions unit & 0 \\
right sub-mammary abscess & -168 \\
spreading cellulitis of the breast & -168 \\
septic & 0 \\
pyrexia of 39.2 & 0 \\
blood pressure of 127/65 & 0 \\
tachycardia of 110 & 0 \\
widespread cellulitis involving the nipple & 0 \\
cellulitis of the sub-mammary area & 0 \\
cellulitis spreading to the axilla & 0 \\
white cell count of 23.8 & 0 \\
neutrophils of 19.9 & 0 \\
CRP of 428 & 0 \\
diagnosed with simple cellulitis & 0 \\
commenced on parenteral antibiotics & 0 \\
fluid resuscitation & 0 \\
intravenous benzyl penicillin & 0 \\
flucloxacillin & 0 \\
condition worsened & 72 \\
surgical opinion sought & 72 \\
evidence of synergistic gangrene & 72 \\
cellulitis spread & 72 \\
areas of growing necrotic ulceration & 72 \\
resuscitation commenced & 72 \\
antibiotic therapy adjusted to imipenem & 72 \\
antibiotic therapy adjusted to clindamycin & 72 \\
taken to theatre within a few hours & 72 \\
partial mastectomy performed & 72 \\
wound left open and packed & 72 \\
returned to theatre for further debridement & 96 \\
vacuum dressing applied & 96 \\
signs of sepsis improved & 120 \\
returned to the ward & 120 \\
blood markers improved & 168 \\
secondary closure performed & 312 \\
discharged & 528 \\
follow-up arranged & 528 \\
\bottomrule
\end{tabular}
\end{tcolorbox}
\end{minipage}

\end{tcolorbox}

\caption{Clinical narrative (left) aligned with temporal T2S2 annotations (right).}
\label{fig:fine-grained-t2s2-annotation}
\end{figure*}

\clearpage
\section{Sensitivity of Event Matching Threshold}
\label{apd:threshold_sweep}

Event matching is performed using cosine distance between PubMedBERT embeddings with threshold $\theta = 0.10$ (Section~\ref{sec:evaluation_methodology_metrics}). Because the event match rate depends directly on $\theta$, we conducted a sensitivity analysis by sweeping $\theta$ from $0.01$ to $0.50$ in increments of $0.01$ and recomputing match rate, concordance, and AULTC for each model.

\begin{figure}[!ht]
    \centering
    \begin{minipage}[t]{0.48\columnwidth}
        \centering
        \includegraphics[width=\linewidth]{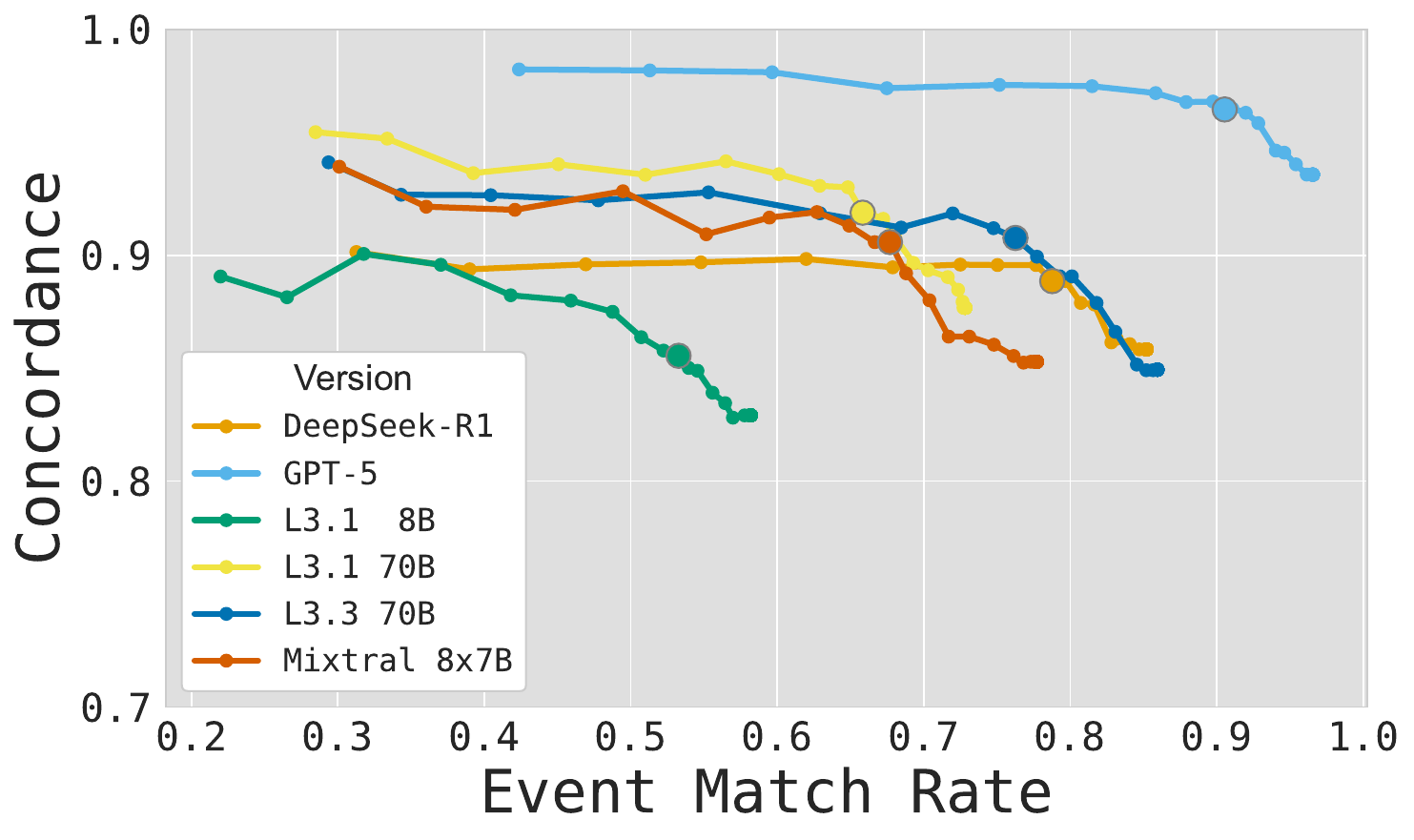}
        \label{fig:threshold_sweep:concordance}
    \end{minipage}\hfill
    \begin{minipage}[t]{0.48\columnwidth}
        \centering
        \includegraphics[width=\linewidth]{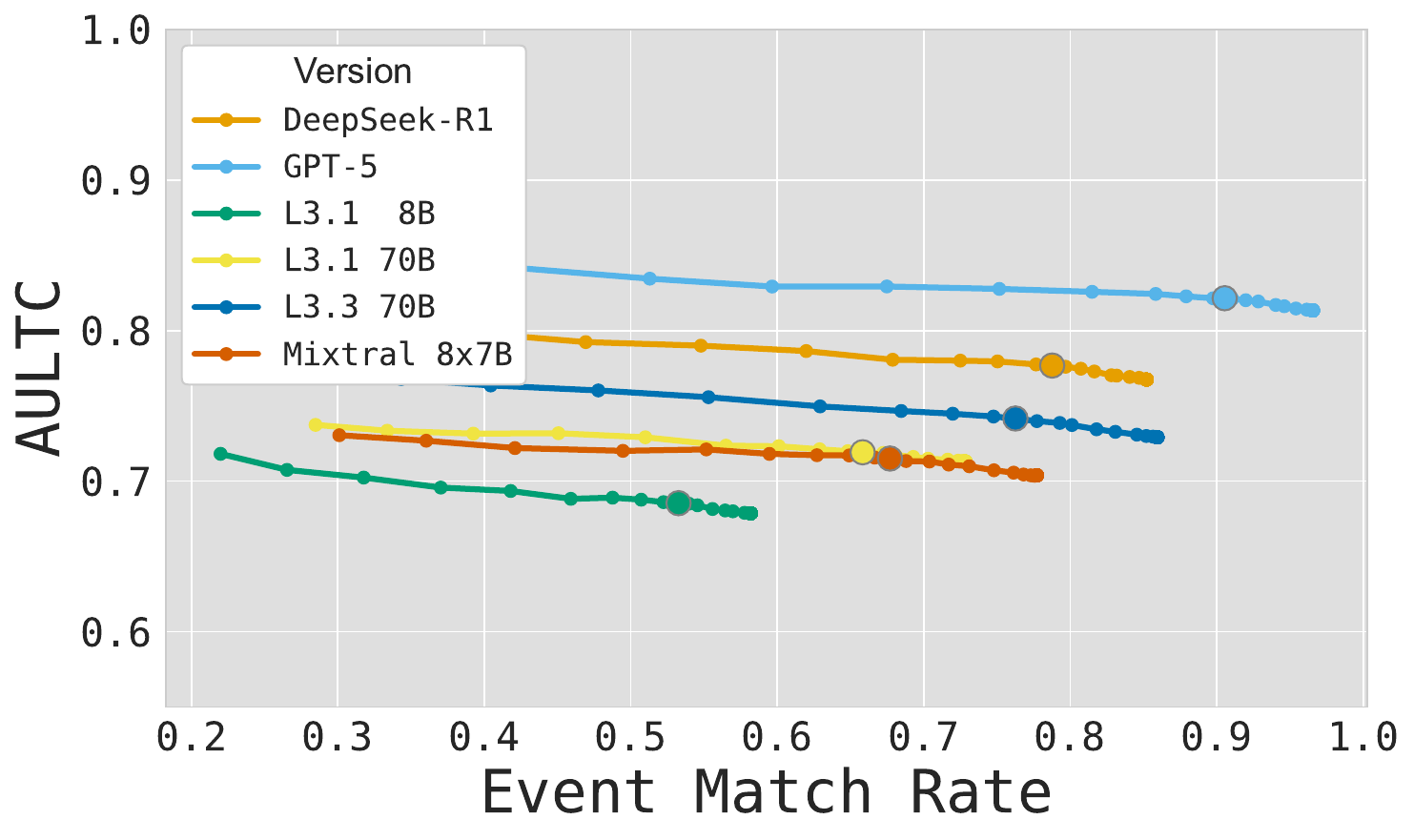}
        \label{fig:threshold_sweep:aultc}
    \end{minipage}
    \caption{Sensitivity of temporal metrics to the cosine distance threshold $\theta$ for the recursive matching heuristic. Left: Concordance versus event match rate obtained by sweeping $\theta$ in increments of $0.01$ over $[0.01, 0.50]$. Right: AULTC versus event match rate under the same sweep. Each point ($\cdot$) corresponds to a distinct threshold value; the solid large circle ($\bullet$) denotes the threshold $\theta = 0.10$ used throughout the paper.}
    \label{fig:threshold_sweep}
\end{figure}

Figure~\ref{fig:threshold_sweep} shows concordance and AULTC as functions of event match rate induced by varying $\theta$ on the \texttt{sepsis-40} dataset. As expected, increasing $\theta$ monotonically increases event match rate. However, beyond a moderate threshold, additional matches correspond to increasingly loose semantic alignments, which degrade temporal metrics. This produces a characteristic trade-off curve: initial increases in match rate are accompanied by minimal degradation in concordance and AULTC, followed by a sharper decline once lower-quality matches are admitted.

Across models, we observe a clear ``elbow'' region near $\theta = 0.10$--$0.15$, where match rate increases modestly while concordance and AULTC remain stable. Beyond this region, particularly past $\theta \approx 0.20$, both concordance and AULTC begin to decline (with concordance drop being more rapid), indicating that additional matches are likely false positive alignments that impair temporal fidelity.

The threshold $\theta = 0.10$ used in the main analysis lies within this elbow region, balancing semantic precision and temporal robustness. Importantly, relative model ordering remains stable across a wide range of thresholds, suggesting that conclusions regarding comparative performance are not artifacts of a particular cosine distance cutoff of the embeddings.
These results support the use of $\theta = 0.10$ as a conservative and temporally faithful operating point for event matching.

\clearpage
\section{Additional Performance Analysis}\label{apd:add-perf}
\begin{figure}[bh!]
\centering
    \begin{minipage}[t]{0.30\textwidth}
    \centering
    \includegraphics[width=\linewidth, page=1]{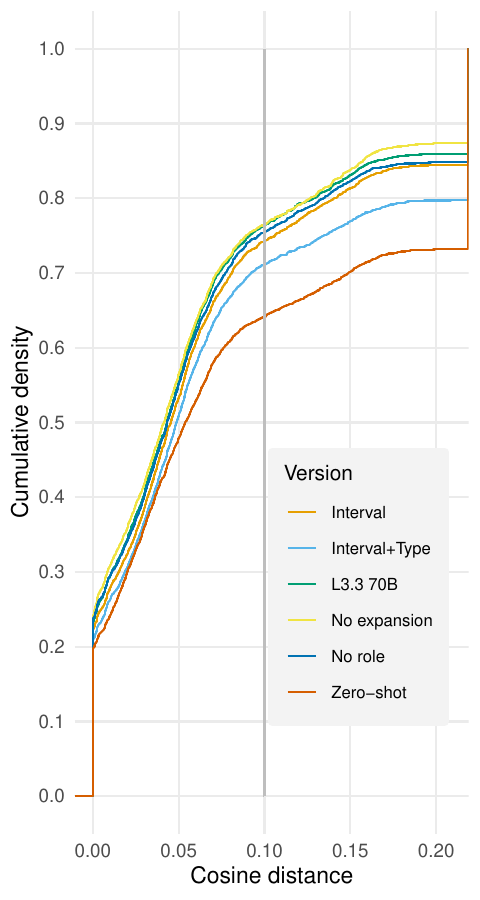}
    \end{minipage}
    \begin{minipage}[t]{0.30\textwidth}
    \centering
    \includegraphics[width=\linewidth, page=2]{figures/main_results_figures/figures_sepsis_40_ablations.pdf}
    \end{minipage}

    \begin{minipage}[t]{0.30\textwidth}
    \centering
    \includegraphics[width=\linewidth, page=3]{figures/main_results_figures/figures_sepsis_40_ablations.pdf}
    \end{minipage}
    \begin{minipage}[t]{0.30\textwidth}
    \centering
    \includegraphics[width=\linewidth, page=4]{figures/main_results_figures/figures_sepsis_40_ablations.pdf}
    \end{minipage}
    \caption{Event match cumulative distribution function (top left) and  concordance box-plots (top right) for the \texttt{sepsis-40} ablations. 
    Time discrepancy from the manual annotation timestamps among matched events, overall (bottom left) and disaggregated by clinician annotator timestamp (time from presentation, bottom right).}
    \label{fig:sep40ablate-time-disc}
\end{figure}

Here we report performance plots for the \texttt{sepsis-40} ablation analysis and the i2m4 dataset (main analysis and ablation), analogous to Figure \ref{fig:time-disc}. Figure \ref{fig:sep40ablate-time-disc} displays the performance of the Llama 3.3 ablations on \texttt{sepsis-40}.  In terms of match rate, the ablations perform within 5 percent of each other except for the 0-shot and Interval plus Typing (Int+Type) methods.  In the temporal assessment figures we see that, for example, the 0-shot method shows promising temporal performance characteristics, but recall that it suffers from much lower event match rates. The Interval plus Typing method (Int+Type) shows a similar pattern.  Requesting the interval alone (Interval) results in worse time discrepancy characteristics as compared to the main approach.  

\begin{figure}[bh!]
\centering
    \begin{minipage}[t]{0.3\textwidth}
    \centering
    \includegraphics[width=\linewidth, page=1]{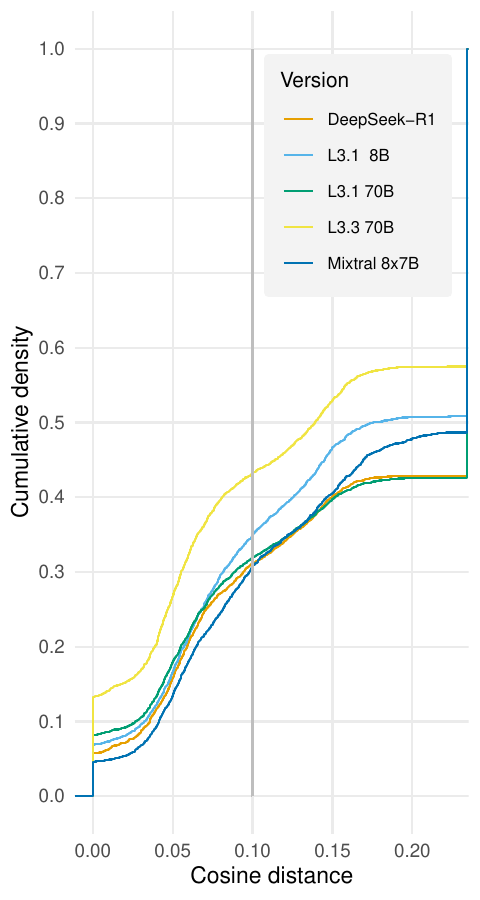}
    \end{minipage}
    \begin{minipage}[t]{0.3\textwidth}
    \centering
    \includegraphics[width=\linewidth, page=2]{figures/main_results_figures/figures_i2m4.pdf}
    \end{minipage}

    \begin{minipage}[t]{0.3\textwidth}
    \centering
    \includegraphics[width=\linewidth, page=3]{figures/main_results_figures/figures_i2m4.pdf}
    \end{minipage}
    \begin{minipage}[t]{0.3\textwidth}
    \centering
    \includegraphics[width=\linewidth, page=4]{figures/main_results_figures/figures_i2m4.pdf}
    \end{minipage}
    \caption{Event match cumulative distribution function (top left) and  concordance box-plots (top right) for the i2m4 dataset. 
    Time discrepancy from the manual annotation timestamps among matched events, overall (bottom left) and disaggregated by clinician annotator timestamp (time from presentation, bottom right).}
    \label{fig:i2m4-time-disc}
\end{figure}

\begin{figure}[ht!]
\centering
    \begin{minipage}[t]{0.3\textwidth}
    \centering
    \includegraphics[width=\linewidth, page=1]{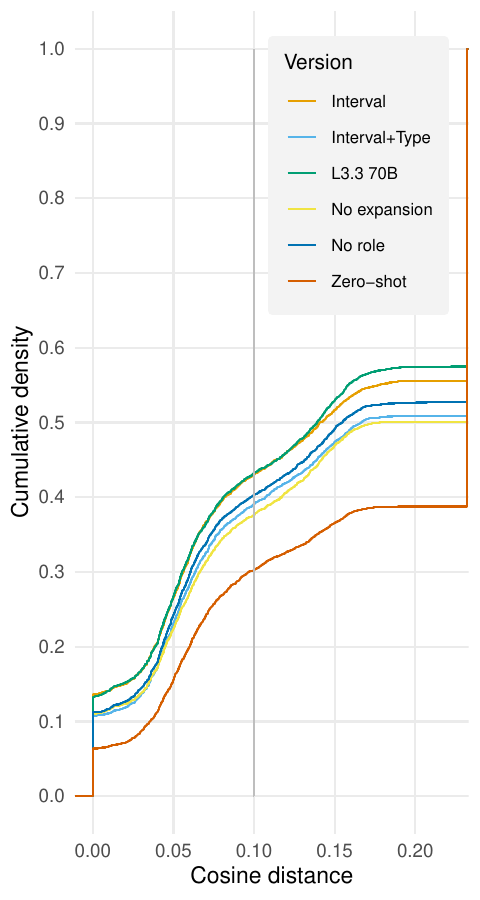}
    \end{minipage}
    \begin{minipage}[t]{0.3\textwidth}
    \centering
    \includegraphics[width=\linewidth, page=2]{figures/main_results_figures/figures_i2m4_ablations.pdf}
    \end{minipage}

    \begin{minipage}[t]{0.3\textwidth}
    \centering
    \includegraphics[width=\linewidth, page=3]{figures/main_results_figures/figures_i2m4_ablations.pdf}
    \end{minipage}
    \begin{minipage}[t]{0.3\textwidth}
    \centering
    \includegraphics[width=\linewidth, page=4]{figures/main_results_figures/figures_i2m4_ablations.pdf}
    \end{minipage}
    \caption{Event match cumulative distribution function (top left) and  concordance box-plots (top right) for the i2m4 ablations. 
    Time discrepancy from the manual annotation timestamps among matched events, overall (bottom left) and disaggregated by clinician annotator timestamp (time from presentation, bottom right).}
    \label{fig:i2m4ablate-time-disc}
\end{figure}

Figures \ref{fig:i2m4-time-disc} and \ref{fig:i2m4ablate-time-disc} displays the performance of the models and the Llama 3.3 ablations on the i2m4 dataset.  The match rates and concordances are visibly lower in the i2m4 dataset than in \texttt{sepsis-40}. Comparing across methods, Llama 3.3 visibly outperforms the Llama 3.1 models in both measures.
Despite the lower match rates and concordances, the time discrepancies in i2m4 were comparatively stronger (AULTC of 0.703). In terms of time discrepancy, Llama 3.1 70B demonstrate smaller time discrepancies, but at the cost of an event match rate of only 30 percent.
In the ablation analysis of the i2m4 dataset, we observe that the Interval technique performs very similarly to the Llama 3.3 version (unlike in the \texttt{sepsis-40} dataset, where it underperforms markedly with respect to time discrepancy).

\clearpage

\section{Additional Corpus Validation Details}
\label{apd:validation_details}

This appendix provides details for three validation analyses summarized in Section~\ref{sec:additional_validation_summary}: the \texttt{sepsis-10} hallucination audit, multi-pass self-consistency analysis, and full-corpus cross-model agreement.
The \texttt{sepsis-10} PMCID composition is provided in Appendix~\ref{apd:subset_ids}.

\subsection{Manual hallucination audit}
\label{apd:hallucination_audit_results}
Because automatically generated corpora can be affected by unsupported or fabricated events, we directly audited the factual grounding of extracted events in 10 sampled PMOA case reports (\texttt{sepsis-10}).
Across DeepSeek-R1 and Llama-3.3-70B-Instruct, this audit covered 1,517 extracted events.
Of these, 1,352 events were grounded automatically by conservative lexical matching, while 165 required manual review because they involved transformations such as abbreviation expansion, conjunction splitting, or semantic paraphrasing.
After manual adjudication of all unresolved cases against the original reports, we identified \textbf{0} hallucinated events (Table~\ref{tab:hallucination_audit}).
This result does not prove that hallucinations never occur in the full corpus, but it provides direct evidence that under the extractive formulation used here, unsupported event fabrication was not observed in the audited subset.

\begin{table*}[ht]
\centering
\small
\caption{Manual hallucination audit on 10 PMOA case reports. Automated lexical grounding used exact substring, ordered token subsequence, or unordered token containment after normalization. Events not grounded automatically were manually reviewed against the source report. Manual review identified 0 hallucinated events.}
\label{tab:hallucination_audit}
\begin{adjustbox}{max width=\textwidth}
\begin{tabular}{lrrrrrr}
\toprule
& \multicolumn{3}{c}{DeepSeek-R1} & \multicolumn{3}{c}{L3.3 70B} \\
\cmidrule(lr){2-4}\cmidrule(lr){5-7}
PMOA case ID & Events & Auto-grounded & Manual review & Events & Auto-grounded & Manual review \\
\midrule
PMC10370906 & 59 & 51 & 8 & 73 & 72 & 1 \\
PMC10556716 & 107 & 95 & 12 & 115 & 110 & 5 \\
PMC10629858 & 77 & 73 & 4 & 63 & 61 & 2 \\
PMC3075162 & 87 & 77 & 10 & 77 & 74 & 3 \\
PMC4778217 & 65 & 63 & 2 & 63 & 60 & 3 \\
PMC6238558 & 70 & 66 & 4 & 71 & 64 & 7 \\
PMC7576661 & 45 & 41 & 4 & 54 & 45 & 9 \\
PMC9136437 & 69 & 59 & 10 & 125 & 124 & 1 \\
PMC9355071 & 58 & 43 & 15 & 57 & 45 & 12 \\
PMC9552858 & 101 & 63 & 38 & 81 & 66 & 15 \\
\midrule
Total & 738 & 631 & 107 & 779 & 721 & 58 \\
\bottomrule
\end{tabular}
\end{adjustbox}
\end{table*}

\paragraph{Hallucination audit interpretation}
The hallucination audit was designed to distinguish unsupported event fabrication from valid clinical transformations.
Events requiring manual review were not automatically considered hallucinations, because T2S2 annotations may legitimately expand abbreviations, split conjunctions, or paraphrase source text into standalone clinical findings.
For example, an extracted event such as ``PET/CT'' may correspond to the source phrase ``positron emission tomography/computed tomography,'' and an extracted event such as ``clinical deterioration'' may correspond to the source phrase ``deteriorated clinically.''
Similarly, medication stop events may require combining a causal statement with a later phrase indicating that multiple drugs were stopped.
After manual review of all unresolved cases in the audited subset, no unsupported events were identified.

\clearpage
\subsection{Multi-pass self-consistency}
\label{apd:self_consistency_results}
We next evaluated whether extraction behavior was stable across stochastic generations.
For each model, Run 1 was treated as the within-model reference and Runs 2--5 were compared against it.
All four evaluated models showed high self-consistency across repeated generations (Table~\ref{tab:self_consistency}).
GPT-5 and Llama 3.3 had particularly high event match rates across runs, while DeepSeek-R1 and O1 also showed stable temporal ordering and AULTC.
These results suggest that the extracted timelines are not highly sensitive to random seed variation and that the pipeline does not exhibit the low-consistency behavior often associated with hallucinated or unstable generations.

\begin{table*}[ht]
\centering
\small
\caption{Multi-pass self-consistency across five random seeds on the 10-report hallucination-audit subset. Run 1 is used as the within-model reference; Runs 2--5 are compared against Run 1.}
\label{tab:self_consistency}
\begin{adjustbox}{max width=\textwidth}
\begin{tabular}{llccc}
\toprule
Model & Run & Match rate & Concordance & AULTC \\
\midrule
DeepSeek-R1 & Run 2 & 0.852 & 0.907 & 0.878 \\
DeepSeek-R1 & Run 3 & 0.836 & 0.890 & 0.835 \\
DeepSeek-R1 & Run 4 & 0.839 & 0.892 & 0.860 \\
DeepSeek-R1 & Run 5 & 0.821 & 0.878 & 0.852 \\
\midrule
L3.3 70B & Run 2 & 0.943 & 1.000 & 0.708 \\
L3.3 70B & Run 3 & 0.937 & 1.000 & 0.753 \\
L3.3 70B & Run 4 & 0.954 & 1.000 & 0.735 \\
L3.3 70B & Run 5 & 0.953 & 1.000 & 0.755 \\
\midrule
O1-preview & Run 2 & 0.838 & 0.942 & 0.828 \\
O1-preview & Run 3 & 0.852 & 0.972 & 0.852 \\
O1-preview & Run 4 & 0.855 & 0.965 & 0.840 \\
O1-preview & Run 5 & 0.847 & 0.941 & 0.847 \\
\midrule
GPT-5 & Run 2 & 0.935 & 0.924 & 0.860 \\
GPT-5 & Run 3 & 0.918 & 0.958 & 0.878 \\
GPT-5 & Run 4 & 0.911 & 0.942 & 0.863 \\
GPT-5 & Run 5 & 0.934 & 0.922 & 0.856 \\
\bottomrule
\end{tabular}
\end{adjustbox}
\end{table*}

\paragraph{Interpretation of self-consistency}
The self-consistency analysis evaluates stability of the extraction process across repeated generations.
High self-consistency does not prove clinical correctness, but low self-consistency would indicate that extracted timelines are sensitive to sampling noise and therefore less reliable as corpus annotations.
The observed consistency across DeepSeek-R1, Llama 3.3, O1-preview, and GPT-5 suggests that the prompt and extraction schema induce stable behavior across repeated runs.

\clearpage
\subsection{Full-corpus cross-model agreement}
\label{apd:full_corpus_agreement_results}
Finally, we evaluated cross-model agreement across all 2,139 T2S2 reports to assess extraction consistency beyond the manually annotated subset.
Treating DeepSeek-R1 as a proxy reference, Llama-3.3-70B-Instruct achieved an event match rate of 0.79, concordance of 0.87, and AULTC of 0.83 across the full corpus (Table~\ref{tab:full_corpus_agreement}).
Because these models differ in architecture and training, this agreement provides a scalable reliability signal that the extracted trajectories are systematic and reproducible across model families.
We emphasize that this analysis is not a substitute for clinical adjudication, but rather a corpus-scale agreement check complementing the gold-standard \texttt{sepsis-40} evaluation, clinician--clinician agreement, hallucination audit, and self-consistency results.

\begin{table}[ht]
\centering
\small
\caption{Full-corpus cross-model agreement across all 2,139 T2S2 reports. DeepSeek-R1 is treated as the proxy reference.}
\label{tab:full_corpus_agreement}
\begin{adjustbox}{max width=\columnwidth}
\begin{tabular}{lccc}
\toprule
Model comparison & Match rate & Concordance & AULTC \\
\midrule
L3.3 70B vs.\ DeepSeek-R1 & 0.79 & 0.87 & 0.83 \\
\bottomrule
\end{tabular}
\end{adjustbox}
\end{table}

\paragraph{Interpretation of full-corpus cross-model agreement}
Full-corpus cross-model agreement complements gold-standard evaluation by estimating whether two independent model families produce similar event--time trajectories across all 2,139 reports.
This analysis is not intended to replace clinician adjudication and cannot identify all shared model errors.
However, strong agreement across model families provides evidence that the extraction behavior is systematic at corpus scale rather than being driven by isolated idiosyncratic outputs from a single model.

\clearpage

\section{Error Analysis}\label{apd:error-analysis}

\subsection{Error Analysis of Event Match Rate}
To investigate the discrepancies in event match rate between the i2m4 and \texttt{sepsis-40} datasets, we analyzed the datasets' structure and examined model performance using Llama 3.3 as a representative model, as the pattern of differences observed across datasets was consistent for all models.

\paragraph{Overview of Clinical Event Identification.}
For the i2m4 dataset, manual annotations by clinicians identified an average of $195\pm94$ clinical events per report across $20$ case reports, compared to $80\pm28$ clinical events per report for \texttt{sepsis-10} across $10$ case reports. 
Llama 3.3 identified an average of $120\pm61$ clinical events for i2m4 and $78\pm24$ clinical events for \texttt{sepsis-10}. These results highlight a systematic bias of under-identification of clinical events in i2m4 compared to \texttt{sepsis-10} by Llama 3.3.

\begin{figure}[b!]
\scriptsize
    \centering
    \includegraphics[width=0.7\linewidth]{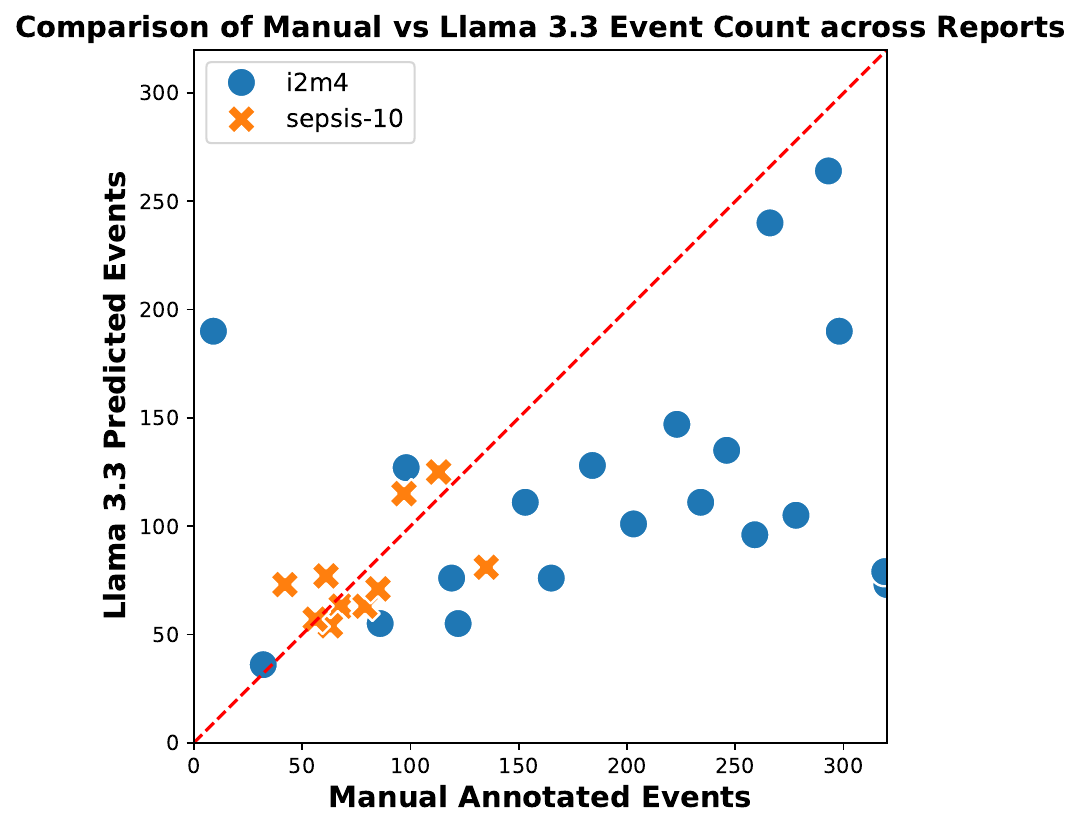}
    \caption{Comparison of Manual Annotation Clinical Event Counts and Llama 3.3 Model Predicted Event Counts: Scatter plot illustrating the variability between manual annotations and Llama 3.3 predictions across reports in two datasets, i2m4 (blue circles) and \texttt{sepsis-10} (orange crosses). The red dashed line indicates the ideal one-to-one correspondence where the number of events per report matches exactly between manual annotations and model predictions.}
    \label{fig:manual_vs_l33_event_count}
\end{figure}

Additionally, the i2m4 dataset contains $2.3\times$ more clinical events per report on average than \texttt{sepsis-10}, as annotated by the clinician, with a much larger variance in event counts across reports (Figure~\ref{fig:manual_vs_l33_event_count}). 
This heterogeneity reflects structural differences in the datasets: i2m4 reports often include a large number of structured entries, whereas \texttt{sepsis-10} reports are only made-up of free text. 
Llama 3.3 over-identified clinical events (i.e., predicted more events than the manual annotators) in $15\%$ of i2m4 reports, while under-identifying events in $85\%$ of i2m4 reports. 
For \texttt{sepsis-10}, the model showed a more balanced distribution, over-identifying and under-identifying events in $50\%$ of reports each (Table~\ref{tab:event_identification_bias}). 
These patterns suggest that the higher density and variability of clinical events in i2m4 pose a greater challenge for Llama 3.3.

\begin{table}[ht]
\setlength{\tabcolsep}{4pt}
\centering
\begin{tabular}{|l|c|c|}
\hline
\textbf{Dataset} & \makecell{\textbf{\% Reports} \\ \textbf{Over-Identified}} & \makecell{\textbf{\% Reports} \\ \textbf{Under-Identified}} \\ \hline
\textbf{i2m4}    & 15\%                                & 85\%                                 \\ \hline
\textbf{\texttt{sepsis-10}} & 50\%                              & 50\%                                 \\ \hline
\end{tabular}
\caption{Proportion of reports where Llama 3.3 over- or under-identified clinical events compared to manual annotations. i2m4 shows a strong bias toward under-identification, whereas \texttt{sepsis-10} has a more balanced distribution.}
\label{tab:event_identification_bias}
\end{table}

\paragraph{Dataset Structure Differences.}
The differences in dataset structure between i2m4 and \texttt{sepsis-10} are also evident in the length of clinical events.
As shown in Table~\ref{tab:character_length}, the manually annotated clinical events in i2m4 are significantly shorter, averaging $10\pm7$ characters, compared to $28\pm16$ characters in \texttt{sepsis-10}. 
This reflects the structured nature of i2m4 reports, which often include tables with abbreviations (e.g., ``K'' for Potassium, ``Na'' for Sodium). 
In contrast, \texttt{sepsis-10} reports are written in free text and provide full terms and descriptions, avoiding the abbreviation issue entirely. 
The absence of abbreviations in \texttt{sepsis-10} reduces variability in event representation and ensures greater alignment between manual annotations and Llama 3.3 predictions.

\begin{table}[ht]
\centering
\begin{tabular}{|l|c|c|}
\hline
\textbf{Dataset} & \textbf{Manual Annotator} & \textbf{Llama 3.3} \\ \hline
\textbf{i2m4}    & $10\pm7 $                              & $17\pm14$                        \\ \hline
\textbf{\texttt{sepsis-10}} & $28\pm16$                            & $21\pm13$                        \\ \hline
\end{tabular}
\caption{Average length of clinical events in characters for both datasets. Manual annotations in i2m4 are significantly shorter due to frequent abbreviations and structured text from MIMIC tables, while Llama 3.3 generates events of comparable length across datasets.}
\label{tab:character_length}
\end{table}

Llama 3.3-generated clinical events are of similar length across datasets, with averages of $17\pm14$ characters for i2m4 and $21\pm13$ characters for \texttt{sepsis-10}. 
This consistency suggests that Llama 3.3 applies similar generative behavior across datasets, often producing expanded forms of abbreviations in i2m4. 
While this generative behavior mitigates variability in Llama 3.3 outputs, it creates a mismatch with the shorter, more concise manual annotations in i2m4. 
This mismatch can then contribute to higher cosine distances between embeddings, lowering the event match rate for i2m4.

\begin{figure}[tb!]
\scriptsize
    \centering
    \includegraphics[width=0.7\linewidth]{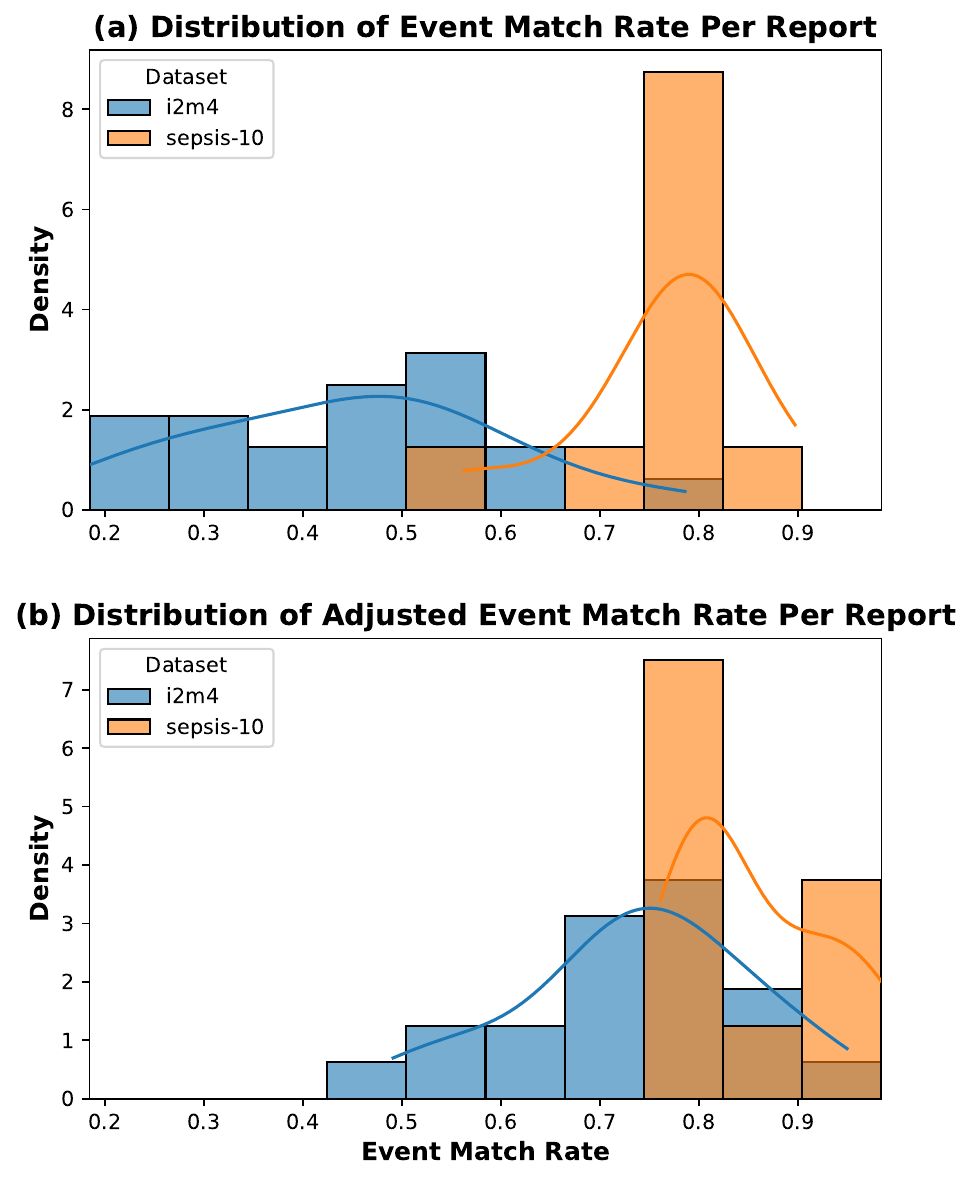}
    \caption{Distributions of Event Match Rates Across Reports Before and After Adjustment: 
    (a) Unadjusted Match Rate: Histogram and density plots of event match rates between manual annotations and Llama 3.3 predictions for reports in i2m4 (blue) and \texttt{sepsis-10} (orange). 
    The lower and more variable match rates in i2m4 highlight the impact of under-identification bias. 
    (b) Adjusted Match Rate: Histogram and density plots showing event match rates after excluding manual clinical events without Llama 3.3 counterparts. 
    The adjustment improves the match rates in i2m4, demonstrating that under-identification is the primary cause of its lower initial match rates compared to \texttt{sepsis-10}.}
    \label{fig:match_rate_distributions}
\end{figure}

\paragraph{Event Match Rate Insights.}
The event match rate, calculated as the proportion of manual clinical events correctly matched by Llama 3.3 predictions based on the sentence transformer embedding cosine distance threshold of 0.1, is significantly lower for i2m4 ($0.404$) than for \texttt{sepsis-10} ($0.753$). 
It should be noted that over-identification of clinical events by Llama 3.3 does not impact this metric, as in calculating the event match rate, we only consider events identified by the manual annotators. 
The observed difference can be attributed to two main factors:

\begin{itemize}
    \item Under-Identification Bias in i2m4: 
    When we adjust for under-identification by excluding manual clinical events for which Llama 3.3 did not identify a counterpart, the average match rate for i2m4 increases from $0.404$ to $0.752$, aligning more closely with \texttt{sepsis-10}'s match rate that increased from $0.753$ to $0.857$. 
    Figure~\ref{fig:match_rate_distributions} illustrates the match rate distributions across reports before and after the adjustment, highlighting the lower and more variable match rates in i2m4 compared to \texttt{sepsis-10}.
    This demonstrates that under-identification is the primary driver of the lower match rate in i2m4. 
    \item Dataset Structure Differences: 
    The structured nature of i2m4 reports introduces variability in the manual annotations due to the frequent use of abbreviations and table-like formats. 
    In contrast, the free-text nature of \texttt{sepsis-10} ensures more consistent and complete representations of clinical events. 
    While Llama 3.3 generates events of similar length across datasets, the expanded text spans of the abbreviations it generates in i2m4 exacerbate the mismatch with the manual annotations, leading to higher cosine distances and a lower event match rate.
\end{itemize}

These findings demonstrate that the lower event match rate for i2m4 is at least partially driven by a systematic under-identification of clinical events by Llama 3.3 in this dataset, compounded by structural differences of reports in i2m4 versus \texttt{sepsis-10}. 

\subsection{Error Analysis of Temporal Predictions}
Across all models, i2m4 exhibits a lower concordance index (c-index) compared to \texttt{sepsis-10}. 
To better understand this discrepancy, we closely examine the performance of Llama 3.3, which shows similar trends in event ordering inconsistencies and a concordance-index drop from 0.93 to 0.75.

In i2m4, we observe higher variance in predicted event timestamps, with some extreme deviations from clinician annotations. To quantify the impact of these outliers, we apply an outlier filtering approach based on the interquartile range ($IQR$). 
The $IQR$ represents the middle 50\% of time differences, defined as the range between the first quartile ($Q1$, $25^{th}$ percentile) and the third quartile ($Q3$, $75^{th}$ percentile). 
Predictions falling outside $Q1 - 1.5 \times IQR$ or $Q3 + 1.5 \times IQR$ are removed, ensuring that event ordering is evaluated on a more stable subset of predictions.

\begin{figure}[tb!]
\scriptsize
    \centering
    \includegraphics[width=0.7\linewidth]{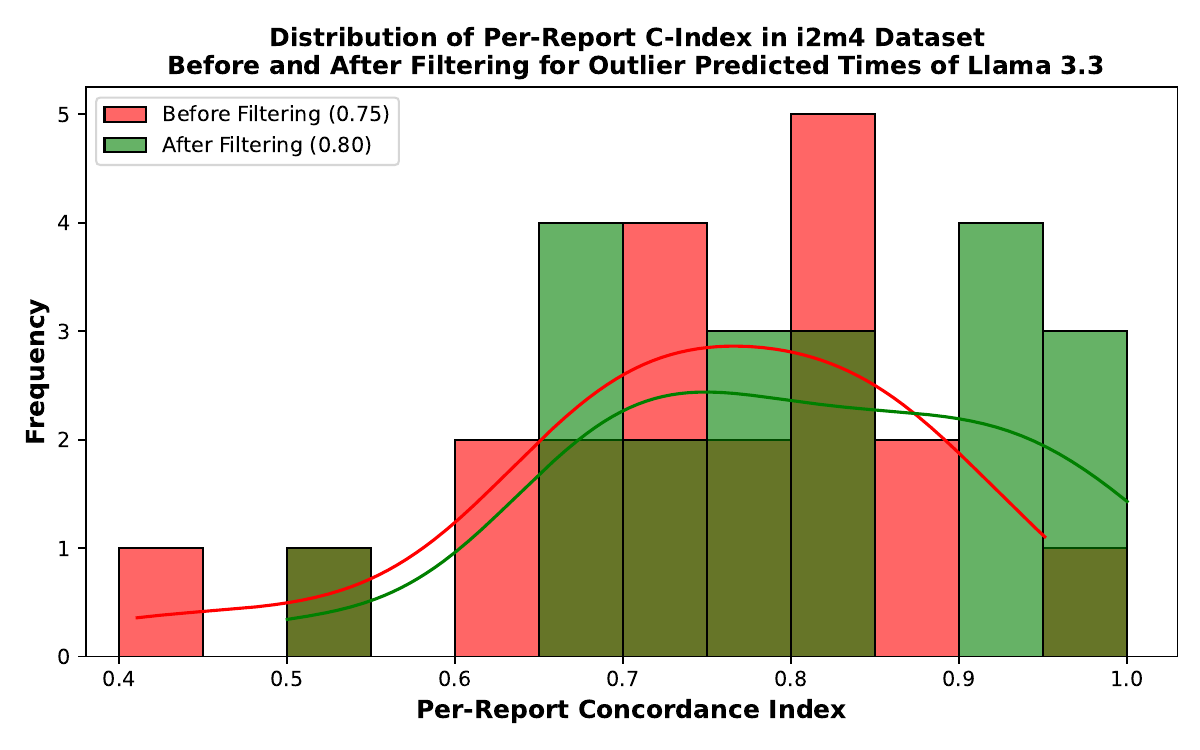}
    \caption{Effect of Outlier Filtering on c-index in i2m4. Distribution of per-report c-index before and after removing extreme time mispredictions using the interquartile range (IQR). Filtering improves the average c-index by reducing the impact of large temporal deviations.}
    \label{fig:cindex_outlier_filtering_l33}
\end{figure}

After filtering, the average c-index for i2m4 improves from 0.75 to 0.80, suggesting that extreme time mispredictions were an important factor in lowering event ranking consistency. 
The shift in per-report c-index distribution after filtering further supports this observation as shown in Figure~\ref{fig:cindex_outlier_filtering_l33}. 
However, despite this improvement, i2m4’s c-index remains lower than \texttt{sepsis-10}, indicating that additional challenges, such as increased annotation complexity or structural differences in event distributions mentioned in the previous section, contribute to its overall lower performance. 

\subsection{Clinical Event Type Categorization Process}
\paragraph{Event Type Category Generation.}
To systematically categorize clinical events identified in case reports from two datasets, \texttt{sepsis-10} and i2m4, we employed a two-step methodology. 
The goal was to classify extracted clinical events into a predefined set of event type categories to use for further analysis of our results. 
Given i2m4's data protection provisions only allowed for local analysis, we used \texttt{sepsis-10} to determine the event type categories, as it was derived from PubMed Open Access case reports.

In the first step, event type categories were generated using a language model. 
Since the appropriate categories were not known a priori, we utilized OpenAI's o3-mini, to analyze all manually annotated clinical events from the \texttt{sepsis-10} dataset and derive six distinct event type categories. 
The model was prompted to examine the extracted events and construct a comprehensive yet non-overlapping set of categories that could effectively organize clinical events within the dataset. 
The full prompt used for this categorization process is as follows:

\begin{tcolorbox}[
  colback=yellow!10,
  colframe=orange!80!black,
  title=Event Type Category Generation Prompt,
  boxrule=2pt
]
\footnotesize
\setlength{\parskip}{1pt}
You are a physician and a machine learning scientist specializing in clinical event extraction. 

Your goal is to categorize extracted clinical events into exactly six distinct and mutually exclusive event categories to ensure clarity and consistency.

Constraints:
Each clinical event must belong to exactly one category.
The categories must be coarse enough to prevent ambiguity.
The sixth category is reserved for ``Other or Unknown.'' The first five categories should comprehensively cover the key types of clinical events.
The categories must be mutually exclusive, meaning an event should unambiguously fit into only one category.

Task:
Identify six well-defined, non-overlapping event categories that best classify the extracted clinical events.
Format the output as a comma-separated table with two columns:
Event Type Category: The name of the category.
Category Number: An integer from 0 to 5, with ``Other or Unknown'' always assigned to 5.

[Category Name], 0

[Category Name], 1

[Category Name], 2

[Category Name], 3

[Category Name], 4

Other or Unknown, 5

Now, based on the extracted clinical events below, generate six non-overlapping event categories and format the output as specified:

...all \texttt{sepsis-40} manual annotated clinical events...

\textless \textbackslash end prefix\textgreater
\end{tcolorbox}

The resulting six event type categories, along with their assigned numeric labels, were:

\begin{itemize}
    \item \textbf{Patient Background and Medical History (0)} -- Events describing patient demographics, prior medical conditions, risk factors, and relevant history before the clinical presentation.
    \item \textbf{Clinical Presentation and Examination Findings (1)} -- Events capturing the patient's initial symptoms, subjective complaints, and observable clinical signs from physical examination.
    \item \textbf{Diagnostic Testing and Results (2)} -- Events related to diagnostic investigations, including laboratory tests, imaging studies, and other diagnostic assessments, along with their results.
    \item \textbf{Clinical Management and Interventions (3)} -- Events detailing treatments and medical interventions, including pharmacological therapies, surgical procedures, and other clinical management strategies.
    \item \textbf{Clinical Course, Outcomes, and Follow-up (4)} -- Events describing the progression of the condition, response to treatment, complications, recovery, prognosis, and post-treatment follow-up.
    \item \textbf{Other or Unknown (5)} -- Events that do not clearly fit into the defined categories or contain insufficient information to be classified.
\end{itemize}

\paragraph{Event Type Category Assignment.}
Once these categories were established, they were used to systematically classify each identified clinical event from both \texttt{sepsis-10} and i2m4. 
The classification was performed separately for clinical events identified through manual annotation and those identified using Llama 3.3. 
To maintain compliance with IRB regulations, all event classifications, including those for i2m4, were executed on a local machine to prevent data transfer to external servers.

For this classification step, we used Llama 3.3 to assign each extracted clinical event to one of the six predefined event type categories. 
The model was provided with a prompt that explicitly listed all event type categories and required it to assign each event to the most appropriate category. 
The classification was performed for both manually annotated events and events identified by Llama 3.3 across the \texttt{sepsis-10} and i2m4 datasets. 
The exact prompt used for this classification task is as follows (the green text includes the few-shot examples provided to the model):

\begin{tcolorbox}[
  colback=yellow!10,
  colframe=orange!80!black,
  title=Event Categorization Prompt,
  boxrule=2pt
]
\footnotesize
\setlength{\parskip}{1pt}
You are a medical professional. 

You are tasked with categorizing clinical events extracted from case reports.

Assign the following clinical event to one of these categories.

The categories are:

Patient Background and Medical History: 0,

Clinical Presentation and Examination Findings: 1,

Diagnostic Testing and Results: 2,

Clinical Management and Interventions: 3,

Clinical Course, Outcomes, and Follow‐up: 4,

Other or Unknown: 5
\\
The categories are defined as follows:

Patient Background and Medical History: (Includes demographic details, prior medical diagnoses, past surgical histories, medication use as part of chronic history, and other baseline background information.),

Clinical Presentation and Examination Findings: (Includes symptoms at presentation, physical exam findings---including vital signs, neurological scores that reflect exam observations---and other immediate clinical observations.),

Diagnostic Testing and Results: (Includes all imaging studies, laboratory tests, diagnostic procedures and their reported findings, and formal diagnostic conclusions reached via workup.),

Clinical Management and Interventions: (Includes all treatments, procedures, medications administered acutely, operations, supportive care measures, and decisions/interventions intended to alter the patient’s condition.),

Clinical Course, Outcomes, and Follow‐up: (Includes statements about change in clinical status, response to treatment, complications, transitions in care, recovery, discharge, and long‐term outcomes.),

Other or Unknown: (For events that do not clearly fit into any of the above five categories.)

\color{darkteal}
For example, here is a list of clinical event text and the corresponding category:

Examples:

``60-year-old female'' → 0

``history of atrial fibrillation'' → 0

``weighed 95 kg'' → 0

``Impaired consciousness'' → 1

``high-grade fever'' → 1

``persistently high-temperature spikes'' → 1

``Head CT'' → 2

``no hepatitis A'' → 2

``stage IV lymphoma'' → 2

``successfully treated with fidaxomicin'' → 3

``shifted to cefepime'' → 3

``intravenous immunoglobulins for 5 days'' → 3

``follow-up evaluations were recommended'' → 4

``transferred to a geriatric medicine unit'' → 4

``Discharge'' → 4

``he'' → 5

``confined, 5

``other symptoms'' → 5
\color{black}

Event: ``$\{\text{event}\_\text{text}\}$''

Respond with only the corresponding integer (0-5) from the list above.
You have to pick only one category for each event. 
If there is no clear category, choose the category 5 that corresponds to ``Other or Unknown'' category.
If there is more than one category, choose the category that you think is most relevant one.
Do NOT include any extra text in your response. Do NOT show your thought process.
Only provide the integer corresponding to the category and nothing else.

Format your response as:

Response: \textless integer \textgreater

\textless \textbackslash end prefix\textgreater
\end{tcolorbox}

\paragraph{Event Type Category Alignment and Match Rate Analysis.}  

To assess the consistency of event type categorization between the two annotators (manual annotator vs. Llama 3.3), we computed the event type category alignment rate for each dataset separately.  
This metric quantifies how often the manually annotated events and the Llama 3.3-extracted events---when determined to be aligned---are also assigned to the same event type category.  

For each case report in the dataset, clinical events extracted by the manual annotator were aligned with those extracted by Llama 3.3 using an iterative matching approach based on the cosine distance of their sentence-transformer embeddings.  
Aligned event pairs, which consist of one event from the manual annotator and one from Llama 3.3, were then categorized independently by Llama 3.3 into one of six predefined event type categories of the previous section.  

The event type category alignment rate was computed by determining, for each aligned event pair, whether both events were assigned to the same category.  
If both events in an aligned pair received the same category label, the event type alignment was considered successful.  
For each dataset, the event type alignment rate was first calculated per case report by dividing the number of successfully aligned event type pairs by the total number of aligned pairs in that case report.  
These per-case alignment rates were then averaged across all case reports in the dataset to obtain the final alignment rate.  
Table \ref{tab:event_category_alignment} reports the mean alignment rate along with the standard deviation for each dataset. 

The event type category alignment rate was higher for the i2m4 dataset ($0.71 \pm 0.05$) compared to the \texttt{sepsis-10} dataset ($0.56 \pm 0.05$).  
We hypothesize that this discrepancy is due to the absence of the ``Other or Unknown'' category in the \texttt{sepsis-10} dataset, which likely introduced ambiguity in category assignments.  
Specifically, since no events in \texttt{sepsis-10} were assigned to the ``Other or Unknown'' category, the model may have redistributed events that would have otherwise fallen into this category among the other five categories.  
This misalignment in categorization likely contributed to the lower agreement rate between the two annotators in the \texttt{sepsis-10} dataset.  

\begin{table}[ht]
\centering
\begin{tabular}{|l|c|}
\hline
\textbf{Dataset} & \textbf{Event Type Alignment} \\ \hline
\textbf{i2m4}    & $0.71\pm0.05 $                       \\ \hline
\textbf{\texttt{sepsis-40}} & $0.56\pm0.05$                      \\ \hline
\end{tabular}
\caption{Mean event type category alignment rate between the manual annotator and Llama 3.3 for each dataset.  
The alignment rate is computed as the proportion of aligned event pairs that were assigned to the same event type category.  
The reported values represent the mean alignment rate across case reports, with standard deviation indicated.}
\label{tab:event_category_alignment}
\end{table}

\begin{figure}[tb!]
\scriptsize
    \centering
    \includegraphics[width=0.9\linewidth]{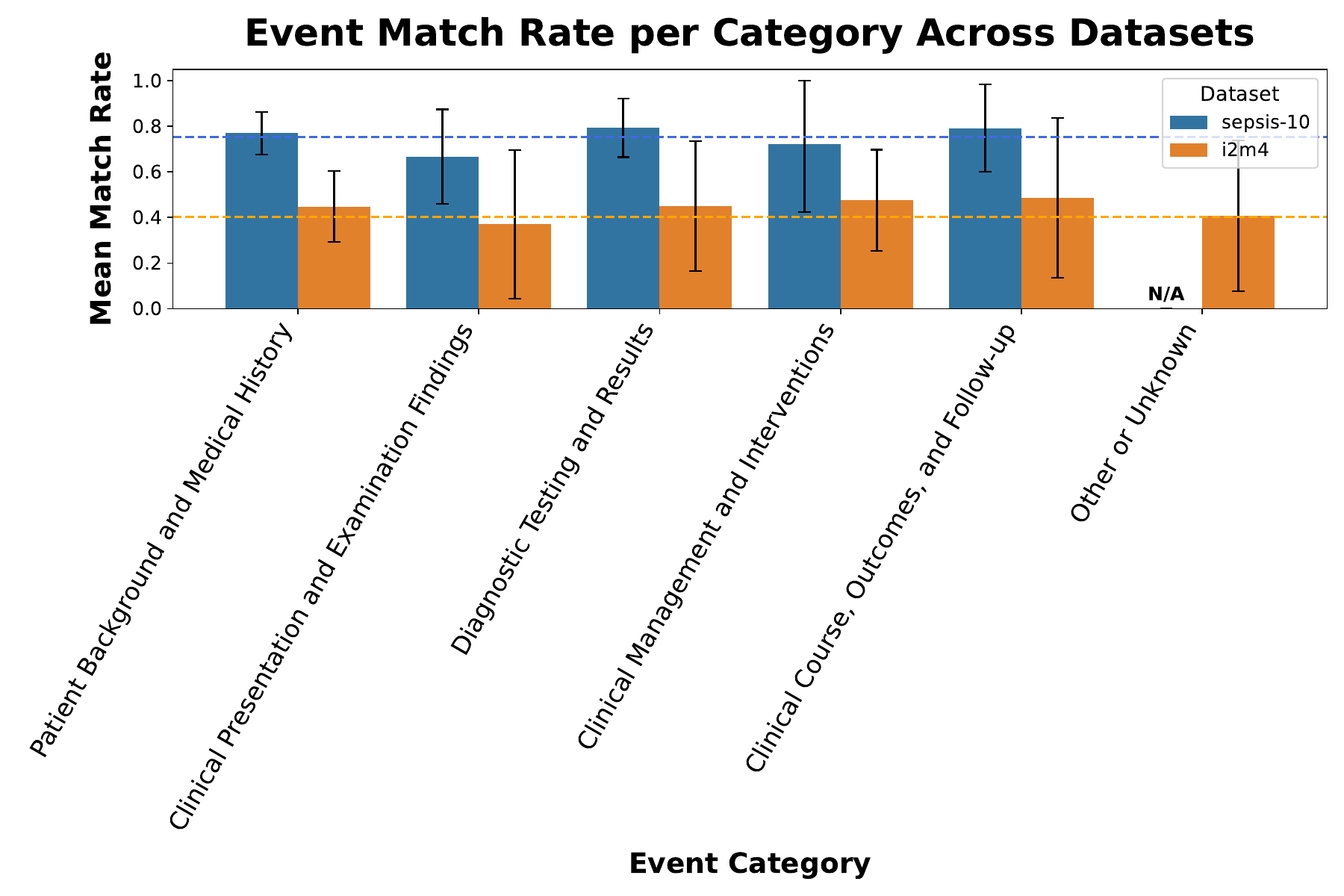}
    \caption{Mean event match rate across predefined event type categories for the i2m4 and \texttt{sepsis-10} datasets.  
    Bars represent the mean match rate for each event type category, calculated as the proportion of aligned event pairs (between the manual annotator and Llama 3.3) with cosine distance less than the matching threshold.  
    Error bars indicate the standard deviation of match rates across case reports.  
    The dashed lines represent the overall mean match rates for each dataset across all categories (From Table~\ref{tab:results}: \textcolor{blue}{\texttt{sepsis-10}}: 0.753 and \textcolor{orange}{i2m4}: 0.404)
    The ``Other or Unknown'' category is absent in \texttt{sepsis-10}, resulting in no match rate for this category in that dataset.}
    \label{fig:event_match_rate_per_category}
\end{figure}

To further investigate the impact of event type categorization on event match rates, we analyzed the match rate separately for each of the six predefined event categories within both datasets.  
Figure \ref{fig:event_match_rate_per_category} displays the mean match rate for each event category, with error bars representing the standard deviation across case reports.  
The dashed horizontal lines indicate the overall mean match rate for all event types within each dataset (as seen in Table~\ref{tab:results}), providing a point of reference for comparison.

From the figure, it is evident that the event match rates exhibit variability across categories but do not show significant deviations that would suggest one category consistently performs better or worse than the others.  
For instance, in both datasets, categories such as ``Diagnostic Testing and Results'' and ``Clinical Course, Outcomes, and Follow up'' tend to achieve higher match rates compared to others, though the differences are not substantial enough to be statistically significant.  
Similarly, the match rates for ``Clinical Management and Interventions'' and ``Patient Background and Medical History'' fall closer to the overall dataset mean, suggesting these categories are neither particularly challenging nor exceptionally easy for alignment.

A notable observation is the ``Other or Unknown'' category, which is entirely absent in the \texttt{sepsis-10} dataset.  
This absence introduces a unique challenge for categorization, as events that might have otherwise been assigned to this category are redistributed among the remaining categories.  
The i2m4 dataset, on the other hand, includes this category, and has a moderate event match rate for the event types in this category.  
This discrepancy could partially suggest that the style, structure, or format of the \texttt{sepsis-10} dataset differs from that of i2m4.  
The absence of the ``Other or Unknown'' category in \texttt{sepsis-10} could indicate a more standardized or narrowly focused structure, where fewer ambiguous or unclassifiable events are present, whereas i2m4 may include a broader range of events requiring such a fallback category.  
This difference in dataset characteristics could contribute to the observed variability in event match rates, as the annotation and alignment processes are inherently influenced by the dataset's structure and complexity.  

Overall, the analysis suggests that event match rates are relatively stable across event categories within each dataset.  
The differences in overall match rates between i2m4 and \texttt{sepsis-10} (and hence \texttt{sepsis-40}) are therefore more likely attributable to dataset-specific characteristics and under-identification bias that was explained in the previous section.
Future work could explore more granular metrics, such as examining the semantic content of events within each category, to better understand the nuances of these event match rate discrepancies. 

\end{document}

%% file: tables/datasets.tex
\begin{table*}[tb!]
\caption{Dataset summary. Bold cells indicate contributed labeling}
\label{tab:dataset-summary}

\centering
\begin{adjustbox}{width=0.8\textwidth}
\begin{tabular}{|p{3.5cm}|p{3.5cm}|p{3.5cm}|p{3.5cm}|}
\hline
\textbf{Corpus}                  & \centering \textbf{i2m4}             & \centering \textbf{\texttt{sepsis-40}}& \textbf{\qquad \texttt{sepsis-100}}       \\
\hline
\textbf{N (unlabeled)}          & \multicolumn{1}{r|}{310;  331,794}             & \multicolumn{1}{r|}{2,139}                     & \multicolumn{1}{r|}{2,139}                      \\ \hline
\textbf{N (labeled)}            & \multicolumn{1}{r|}{20}                         & \multicolumn{1}{r|}{40}                       & \multicolumn{1}{r|}{100}                       \\ \hline
\textbf{Source}                 & i2b2; MIMIC IV           & PMOA       & PMOA       \\ \hline
\textbf{Note type}              & discharge summary        & case report              & case report               \\ \hline
\textbf{Labeler}                & \textbf{clinicians}               & \textbf{clinicians} (Gold)& \textbf{GPT-5} (Bronze)\\ \hline
\textbf{Phenotyper}             & \textbf{clinicians}      & \textbf{clinicians}      & \textbf{QwQ-32B-Preview}             \\ \hline
\textbf{Methods}                &                & GPT-5& GPT-5\\ 
                                & DeepSeek-R1& DeepSeek-R1& DeepSeek-R1\\ 
                                & L3.3 70B Instruct                  & L3.3 70B Instruct                & L3.3 70B Instruct                 \\ 
                                & L3.1 70B Instruct                 & L3.1 70B Instruct                & L3.1 70B Instruct   \\ 
                                & Mixtral 8x7B & Mixtral 8x7B                & Mixtral 8x7B                   \\ 
                                & \quad Instruct v0.1 & \quad Instruct v0.1                & \quad Instruct v0.1                   \\ 
                                & L3.1 8B Instruct             & L3.1 8B Instruct                & L3.1 8B Instruct                  \\ \hline
\parbox{3cm}{\textbf{Ablations}} & No role                  & No role                  & No role                   \\ 
Base: & No conjunction & No conjunction & No conjunction
                                \\ 
L3.3 70B Instruct                        & \quad instruction &  \quad instruction & \quad instruction \\
                                
                                & No few-shot prompting    & No few-shot prompting    & No few-shot prompting     \\ \hline
\textbf{Sensitivity analyses}   & With upper bound         & With upper bound         & With upper bound          \\ 
                                & With upper bound + 
                                                         & With upper bound +  
                                                                                      & With upper bound + \\
                                & \quad i2b2 event typing & \quad i2b2 event typing & \quad i2b2 event typing \\
                                                                                      \hline
\end{tabular}
\end{adjustbox}
\end{table*}

%% file: tables/mainresults3.tex
\begin{table*}[tb!]
\small
\caption{Comparison of event match rate (Event), median concordance ($c$), median absolute error (MAE), and area under the log-time curve (AULTC) across LLMs for i2m4, \texttt{sepsis-40}, and \texttt{sepsis-100} datasets.}
\label{tab:results}
\footnotesize
\centering

\begin{tabular}{l>{\centering\arraybackslash}p{0.5cm}>{\centering\arraybackslash}p{0.65cm}>{\centering\arraybackslash}p{0.6cm}>{\centering\arraybackslash}p{0.9cm}>{\centering\arraybackslash}p{0.5cm}>{\centering\arraybackslash}p{0.65cm}>{\centering\arraybackslash}p{0.6cm}>{\centering\arraybackslash}p{0.9cm}>{\centering\arraybackslash}p{0.5cm}>{\centering\arraybackslash}p{0.65cm}>{\centering\arraybackslash}p{0.6cm}>{\centering\arraybackslash}p{0.9cm}}
\toprule

& \multicolumn{4}{c}{i2m4}
& \multicolumn{4}{c}{\texttt{sepsis-40}}
& \multicolumn{4}{c}{\texttt{sepsis-100}} \\

\cmidrule(lr){2-5}
\cmidrule(lr){6-9}
\cmidrule(lr){10-13}

\textbf{Model}
& \scriptsize Event
& \scriptsize c
& \scriptsize MAE
& \scriptsize AULTC
& \scriptsize Event
& \scriptsize c
& \scriptsize MAE
& \scriptsize AULTC
& \scriptsize Event
& \scriptsize c
& \scriptsize MAE
& \scriptsize AULTC \\

\midrule

GPT-5
& -- & -- & -- & --
& {\bftab 0.93} & {\bftab 0.965} & {\bftab 4} & {\bftab 0.822}
& -- & -- & -- & -- \\

DeepSeek-R1
& 0.31 & 0.755 & 13 & 0.786
& 0.79 & 0.892 & 16 & 0.781
& {\bftab 0.77} & 0.879 & {\bftab 2} & {\bftab 0.812} \\

L3.3 70B
& {\bftab 0.43} & {\bftab 0.768} & {\bftab 7} & {\bftab 0.805}
& 0.76 & 0.908 & 24 & 0.742
& 0.69 & {\bftab 0.920} & 7 & 0.796 \\

L3.1 70B
& 0.32 & 0.747 & 8 & 0.804
& 0.66 & 0.931 & 30 & 0.737
& 0.67 & 0.914 & 6 & 0.794 \\

Mixtral 8x7B
& 0.31 & 0.745 & 21 & 0.773
& 0.68 & 0.906 & 48 & 0.710
& 0.63 & 0.875 & 24 & 0.764 \\

L3.1 8B
& 0.35 & 0.727 & 26 & 0.738
& 0.53 & 0.856 & 84 & 0.685
& 0.66 & 0.772 & 50 & 0.716 \\

BERT 5-CV
& -- & 0.704 & 34 & 0.520
& -- & -- & -- & --
& -- & -- & -- & -- \\

\bottomrule
\end{tabular}
\end{table*}

%% file: tables/sensitivities3.tex
\begin{table*}[tb!]
\scriptsize
\caption{Comparison of event match rate (Event), median concordance ($c$), median absolute error (MAE), and area under the log-time curve (AULTC) across Llama-3.3-70B-Instruct prompt variants: no role-playing (no role), no conjunction expansion (no expansion), zero-shot, interval, and interval with i2b2 event typing (Interval + Type), for i2m4, \texttt{sepsis-40}, and \texttt{sepsis-100}.}
\label{tab:sens}
\footnotesize
\centering

\begin{tabular}{l>{\centering\arraybackslash}p{0.5cm}>{\centering\arraybackslash}p{0.65cm}>{\centering\arraybackslash}p{0.6cm}>{\centering\arraybackslash}p{0.9cm}>{\centering\arraybackslash}p{0.5cm}>{\centering\arraybackslash}p{0.65cm}>{\centering\arraybackslash}p{0.6cm}>{\centering\arraybackslash}p{0.9cm}>{\centering\arraybackslash}p{0.5cm}>{\centering\arraybackslash}p{0.65cm}>{\centering\arraybackslash}p{0.6cm}>{\centering\arraybackslash}p{0.9cm}}
\toprule

& \multicolumn{4}{c}{i2m4}
& \multicolumn{4}{c}{\texttt{sepsis-40}}
& \multicolumn{4}{c}{\texttt{sepsis-100}} \\

\cmidrule(lr){2-5}
\cmidrule(lr){6-9}
\cmidrule(lr){10-13}

\textbf{Model}
& \scriptsize Event
& \scriptsize c
& \scriptsize MAE
& \scriptsize AULTC
& \scriptsize Event
& \scriptsize c
& \scriptsize MAE
& \scriptsize AULTC
& \scriptsize Event
& \scriptsize c
& \scriptsize MAE
& \scriptsize AULTC \\

\midrule

L3.3 70B
& {\bftab 0.43} & 0.768 & {\bftab 7} & {\bftab 0.805}
& {\bftab 0.76} & 0.908 & 24 & 0.742
& 0.69 & 0.920 & 7 & {\bftab 0.796} \\
\\

\textbf{Ablations} \\
No role
& 0.40 & 0.734 & 8 & 0.800
& 0.75 & 0.920 & 24 & 0.749
& 0.69 & 0.908 & 8 & 0.792 \\

No expansion
& 0.38 & 0.768 & 8 & 0.797
& 0.76 & 0.909 & 24 & {\bftab 0.774}
& 0.69 & 0.923 & 7 & 0.795 \\

Zero-shot
& 0.30 & 0.737 & 9 & 0.794
& 0.64 & {\bftab 0.940} & 24 & 0.746
& 0.59 & {\bftab 0.943} & 12 & 0.786 \\
\\

\textbf{Augments} \\
Interval
& 0.43 & 0.740 & 8 & 0.798
& 0.74 & 0.924 & 24 & 0.737
& {\bftab 0.70} & 0.902 & 10 & 0.789 \\

Interval+Type
& 0.39 & {\bftab 0.775} & 8 & 0.803
& 0.71 & 0.911 & 28 & 0.730
& 0.68 & 0.906 & {\bftab 6} & 0.794 \\

\bottomrule
\end{tabular}
\end{table*}

%% file: sections/aultc.tex
\section{Log-Time Cumulative Distribution Function}
\label{apd:ltcdf}

Recall the log-time cumulative distribution function is given as follows:

\[
   F(x) = \frac{1}{k} \sum_{i=1}^k \mathbf{1}_{\{x_{(i)} \leq x\}}
\]
where $\mathbf{1}$ is the indicator function.

We compute the AULTC as the area under $F(x)$ from $x = 0$ to $x = \log(1 + S_{\text{max}})$, normalized by $\log(1 + S_{\text{max}})$:
\begin{align*}
\text{AULTC} &= \frac{1}{\log(1 + S_{\text{max}})} \Bigg[ \sum_{i=1}^k (x_{(i)} - x_{(i-1)}) \frac{i}{k} + (\log(1 + S_{\text{max}}) - x_{(k)}) \times 1 \Bigg]
\end{align*}
where $x_{(0)} = 0$. With this definition, AULTC = 1 indicates that discrepancies are zero (perfect recovery), resulting in maximum area $\log(1 + S_{\text{max}})$, and AULTC = 0 indicates that all discrepancies exceed $S_{\text{max}}$, yielding zero area. 
The code for the AULTC calculation is given in the function \texttt{ecdf\textunderscore auc2(\dots)} in the Supplementary Materials.

\begin{remark}
The time unit and cutoff $S_\text{max}$ affect the AULTC calculation, so they must be specified when reporting the AULTC.
\end{remark}
Unless otherwise stated, we set $S_{\max}=8760$ hours, corresponding to one year.
This upper limit was chosen because it captures clinically meaningful discrepancies from hours to months while preventing very long-horizon case-history events from dominating the metric.

Because of the $\log(1 + \cdot)$ transformation (rather than the $\log(\cdot)$ transformation which is undefined for zero time-error discrepancy), the discrepancies are non-linearly shifted in the log scale.  The non-linear shift adjusts the relative widths of the step function (particularly for small discrepancies) changing the area calculation.  The cutoff $S_\text{max}$ affects the normalization factor.  Therefore, these values should be chosen based on practicalities for the application.  In our case, we chose hours at the time unit because the only sub-hour descriptions reported were several Apgar scores at the minute-level.

\begin{remark}
    The average log-time discrepancy is non-convex.
\end{remark}
This can be seen from the observation that the $\log(1+\cdot)$ function is ``spikier'' than the L1 function. More formally, the log-time discrepancy has a higher curvature around zero error compared to the L1 loss. Consider the second derivatives of the loss functions with respect to the discrepancies $s = |t^p - t^r|$.  For L1 loss $L_1(s) = s$, the second derivative $\frac{d^2 L_1}{ds^2} = 0$ for $s > 0$.
For the log-time loss, $L_{log}(s) = \log(1 + s)$, the second derivative is $\frac{d^2 L_{log}}{ds^2} = -\frac{1}{(1 + s)^2}$. At $s = 0$, the second derivative is $-1$, while the second derivative of the L1 loss is $0$.

To provide a simple counterexample, define the average log-time discrepancy is given by:
\[
L(t^p) = \frac{1}{k} \sum_{i=1}^k \log(1 + \min(|t^p_i - t^r_i|, S_{\text{max}}))
\]
To prove that this function is non-convex, we need to show that there exist $t^{p1}, t^{p2}$ and $\lambda \in (0,1)$ such that
\[
L(\lambda t^{p1} + (1-\lambda) t^{p2}) > \lambda L(t^{p1}) + (1-\lambda) L(t^{p2})
\]
Consider the case where $k=1$, $t^r = 0$, and $S_{\text{max}} = 2$. The loss function is $L(t^p) = \log(1 + \min(|t^p|, 2))$.
Let $t^{p1} = 0$ and $t^{p2} = 3$. Choose $\lambda = 0.5$.
Then $\lambda t^{p1} + (1-\lambda) t^{p2} = 0.5 \times 0 + 0.5 \times 3 = 1.5$.
Now we evaluate the loss function at these points:
\begin{align*}
L(t^{p1}) &= L(0) = \log(1 + \min(|0|, 2)) = 0 \\
L(t^{p2}) &= L(3) = \log(1 + \min(|3|, 2)) = \\
    & \qquad \log(3) \\
L(\lambda t^{p1} + (1-\lambda) t^{p2}) &= L(1.5) = \log(2.5)
\end{align*}
Now we check the convexity condition:
\[
L(1.5) \leq 0.5 L(0) + 0.5 L(3)
\]
\[
\log(2.5) \leq 0.5 \times 0 + 0.5 \times \log(3)
\]
\[
\log(2.5) \leq \log(\sqrt{3})
\]
Since $2.5 > \sqrt{3} \approx 1.732$, we have $\log(2.5) > \log(\sqrt{3})$.
Therefore,
\[
L(\lambda t^{p1} + (1-\lambda) t^{p2}) > \lambda L(t^{p1}) + (1-\lambda) L(t^{p2})
\]
This proves that the average log-time discrepancy is non-convex.
This remark illustrates the loss form, in case it is considered for optimization/model training rather than for assessment purposes.